\documentclass[10pt,twocolumn,letterpaper]{article}

\usepackage[pagenumbers]{cvpr} % To force page numbers, e.g. for an arXiv version

\usepackage{graphicx}
\usepackage{amsmath}
\usepackage{amssymb}
\usepackage{booktabs}
\usepackage[dvipsnames]{xcolor}
\usepackage{numprint}
\npdecimalsign{.}
\nprounddigits{2}

\usepackage[pagebackref,breaklinks,colorlinks]{hyperref}

\usepackage[capitalize]{cleveref}
\crefname{section}{Sec.}{Secs.}
\Crefname{section}{Section}{Sections}
\Crefname{table}{Table}{Tables}
\crefname{table}{Tab.}{Tabs.}

\def\cvprPaperID{36} % *** Enter the CVPR Paper ID here
\def\confName{CVPR}
\def\confYear{2023}

\colorlet{mylinkcolor}{violet}
\colorlet{mycitecolor}{YellowOrange}
\colorlet{myurlcolor}{Aquamarine}

\begin{document}

\makeatletter
\newcommand\thefontsize[1]{{#1 The current font size is: \f@size pt\par}}
\makeatother


\title{\vspace{-0.7cm}Photorealistic and Identity-Preserving Image-Based Emotion Manipulation with Latent Diffusion Models}

\author{Ioannis Pikoulis,\quad Panagiotis P. Filntisis,\quad Petros Maragos \\
School of ECE, National Technical University of Athens, Greece \\
{\tt\small pikoulis.giannis@gmail.com,\quad filby@central.ntua.gr,\quad maragos@cs.ntua.gr} \\
% \and
}
% \maketitle

% \begin{figure}[t!]
%     \centering
%     \begin{minipage}{0.95\textwidth}
%         \centering
%         \includegraphics[width=1.0\textwidth]{img/cover.jpg} 
%     \end{minipage}%
%     \caption{}
%     \label{fig:cover}
% \end{figure}

\twocolumn[{%
\renewcommand\twocolumn[1][]{#1}%
\maketitle
\begin{center}
\vspace{-0.7cm}
    \centering
    \captionsetup{type=figure}
    \includegraphics[width=0.9\textwidth]{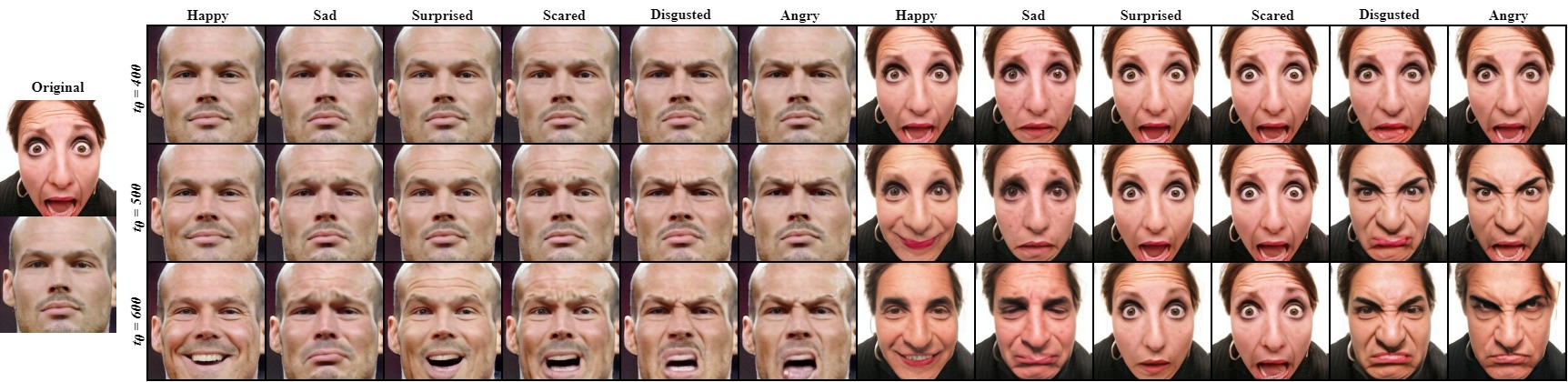}
    \captionof{figure}{Curated examples of LDM-based emotion manipulation on the AffectNet \cite{mollahosseini2017affectnet} validation set, using $T_{\textrm{DDIM}}=40$ steps, $\eta=0$, \textit{c.f.g} scale $\gamma=3.0$ and variable editing strength $t_{0}\in\{400,500,600\}$. Through the proposed method we are able to manipulate facial expressions of “in-the-wild” images while preserving realism, but also control emotion intensity and expression variations.}
    \label{fig:cover}
\end{center}%
}]

%%%%%%%%% ABSTRACT
\begin{abstract}
\vspace{-1.5ex}
    In this paper, we investigate the emotion manipulation capabilities of diffusion models with “in-the-wild” images, a rather unexplored application area relative to the vast and rapidly growing literature for image-to-image translation tasks. Our proposed method encapsulates several pieces of prior work, with the most important being Latent Diffusion models and text-driven manipulation with CLIP latents. We conduct extensive qualitative and quantitative evaluations on AffectNet, demonstrating the superiority of our approach in terms of image quality and realism, while achieving competitive results relative to emotion translation compared to a variety of GAN-based counterparts. Code is released as a publicly available \href{https://github.com/GiannisPikoulis/dsml-thesis/tree/master/face_reenactment}{repo}.
\end{abstract}

\vspace{-3.5ex}
%%%%%%%%% BODY TEXT
\section{Introduction}
\label{sec:intro}
As generative adversarial networks (GANs) \cite{goodfellow2014generative} have proven capable of generating high-quality samples, they managed to draw a lot of attention in the last 10 years. But recently, even more potent generative methods, like diffusion models (DMs) \cite{sohl2015deep, ho2020denoising} have emerged, posing a threat to the dominance of GANs in the production of synthetic data. DMs are quickly finding use in both low-level and high-level vision tasks because of their incredible generative capabilities, including but not limited to image denoising \cite{ho2020denoising, nichol2021improved}, inpainting \cite{esser2021imagebart}, image super-resolution \cite{lugmayr2022repaint, li2022srdiff}, semantic segmentation \cite{baranchuk2022labelefficient, graikos2022diffusion, wolleb2022diffusion}, semantic image synthesis \cite{rombach2022high} and image-to-image translation \cite{chandramouli2022ldedit, kim2022diffusionclip, zhao2022egsde, saharia2022palette}.

In this paper we explicitly focus on image-to-image translation and more specifically on \textit{facial expression manipulation} (also known as face reenactment) on the basis of “in-the-wild” images. Photo-realistic face reenactment can be used for entertainment purposes, human-computer interactions, and facial animations, among other things. This area has been attracting considerable attention both from academic and industrial research communities and has produced stunning outcomes that broaden the scope of inventive image editing, and content creation.

%-------------------------------------------------------------------------
\vspace{-0.15cm}
\section{Related Work}
\subsection{GANs}
The development of GANs spawned an expanding field of study in the area of image and video synthesis. A conditional generator is used in the vast majority of works, meaning that, in the context of emotion manipulation, the synthesized image is dependent on either another image, a class label or any kind of expression-specific information (\eg AU activations). This makes it possible to translate images between various domains (\ie, image-to-image translation) while maintaining the original image's content, even when training on unpaired training data. The multi-domain StarGAN \cite{choi2018stargan,choi2020stargan} framework illustrated how facial expressions may be altered in images by translating them in accordance with the provided semantic label (\eg happy, angry, \etc). ExprGAN \cite{ding2018exprgan} employed \textit{intensity} while in \cite{lindt2019facial} the \textit{valence-arousal} space was utilized for continuous emotion conditioning. By employing a dataset annotated with only categorical labels, the proposed approach of GANmut \cite{d2021ganmut} provided a way to obtain a 2D interpretable emotion-conditional label system. Notable implementations also include GANimation \cite{pumarola2018ganimation}, FACEGAN \cite{tripathy2021facegan} and ICface \cite{tripathy2020icface} which rather opt for emotion conditioning based on facial landmarks, AU intensities.

\subsection{Diffusion Models}
A forward and reverse diffusion process make up the type of latent variable models known as denoising diffusion probabilistic models \cite{sohl2015deep, ho2020denoising} (DDPM). While progressively sampling the latent variables $\mathbf{x}_{t}$, noise is gradually added to the data in the forward process for $t\in[T]$, forming a Markov chain. The forward process's individual steps are all Gaussian, \ie $q(\mathbf{x}_{t}|\mathbf{x}_{t-1}):=\mathcal{N}(\mathbf{x}_{t};\sqrt{1-\beta_{t}}\mathbf{x}_{t-1},\beta_{t}\mathbf{I})$, where $\{\beta_{t}\}^{T}_{t=0}$ denotes a fixed or learned variance schedule. The forward noising process is summarized as follows:
\begin{equation}\label{ddpm_diffusion}
    \mathbf{x}_{t}=\sqrt{\bar{a}_{t}}\mathbf{x}_{0}+\sqrt{1-\bar{a}_{t}}\bm{\epsilon},\quad \bm{\epsilon}\sim\mathcal{N}(\mathbf{0},\mathbf{I})
\end{equation}
where $\bar{a}_{s}=\prod_{s=1}^{t}a_{s}$ and $a_{s}:=1-\beta_{s}$. The reverse, denoising process can also be parameterized in the form of Gaussian transitions $p_{\bm{\theta}}(\mathbf{x}_{t-1}|\mathbf{x}_{t}):=\mathcal{N}(\mathbf{x}_{t-1};\mathbf{\mu}_{\bm{\theta}}(\mathbf{x}_{t},t),\sigma_{t}\mathbf{I})$, where $\mathbf{\mu}_{\bm{\theta}}(\mathbf{x}_{t},t)$ can be decomposed into a linear combination of the noisy input $\mathbf{x}_{t}$ and a noise approximation network $\bm{\epsilon}_{\bm{\theta}}(\mathbf{x}_{t},t)$. Diffusion model training simplifies to the following optimization objective:
\begin{equation}\label{eq:L_simple}
\min_{\bm{\theta}}\mathbb{E}_{\mathbf{x}_{0},\bm{\epsilon},t\sim{U(1,T)}}\lVert\bm{\epsilon}-\bm{\epsilon}_{\bm{\theta}}(\sqrt{\bar{a}_{t}}\mathbf{x}_{0}+\sqrt{1-\bar{a}_{t}}\bm{\epsilon},t)\rVert_{2}^{2}
\end{equation}
Inferring $\mathbf{x}_{t-1}\sim{p_{\bm{\theta}}(\mathbf{x}_{t-1}|\mathbf{x}_{t})}$ can be performed through ancestral sampling:
\begin{equation}\label{eq:ddpm_sampling}
    \mathbf{x}_{t-1}=\frac{1}{\sqrt{a_{t}}}\bigg(\mathbf{x}_{t}-\frac{1-a_{t}}{\sqrt{1-\bar{a}_{t}}}\bm{\epsilon}_{\bm{\theta}}(\mathbf{x}_{t},t)\bigg)+\sigma_{t}\bm{\epsilon}
\end{equation}
where $\bm{\epsilon}\sim\mathcal{N}(\mathbf{0},\mathbf{I})$. 

Song \etal \cite{songddim2021} proposed denoising diffusion implicit models (DDIM), equipped with a non-Markovian diffusion process and jump-step sampling. DDIM shares the same forward marginals with DDPM, but differs in terms of the sampling process:
\begin{equation}\label{eq:ddim_sampling}
    \resizebox{.9\hsize}{!}{$\mathbf{x}_{t-1}=\sqrt{\bar{a}_{t-1}}\mathbf{f}(\mathbf{x}_{t},t)+\sqrt{1-\bar{a}_{t-1}-\sigma_{t}^{2}}\bm{\epsilon}_{\bm{\theta}}(\mathbf{x}_{t},t)+\sigma_{t}\bm{\epsilon}$}
\end{equation}
where $\mathbf{f}_{\bm{\theta}}$ denotes the prediction of the denoised observation $\mathbf{x}_{0}$ at $t$, given $\mathbf{x}_{t}$ and $\bm{\epsilon}_{\bm{\theta}}(\mathbf{x}_{t},t)$:
\begin{equation}\label{eq:f_theta}
    \mathbf{f}_{\bm{\theta}}(\mathbf{x}_{t},t):=\frac{\mathbf{x}_{t}-\sqrt{1-\bar{a}_{t}}\bm{\epsilon}_{\bm{\theta}}(\mathbf{x}_{t},t)}{\sqrt{\bar{a}_{t}}}
\end{equation}
Considering accelerated generation with $\mathbf{\tau}$ being an increasing subsequence of $[1,\dots,T]$, the variance hyperparameter is often parameterized as $\sigma_{\tau_{i}}(\eta)=\eta\sqrt{(1-\bar{a}_{\tau_{i-1}})/(1-\bar{a}_{\tau_{i}})}\sqrt{1-\bar{a}_{\tau_{i}}/\bar{a}_{\tau_{i-1}}}$. 
Especially when $\eta=0$, both the forward and reverse processes become deterministic, enabling fast inversion of the noised input variables. In this case, the resulting model becomes an implicit probabilistic model \cite{mohamed2016learning}.

In an attempt to increase speed and computational efficiency, latent diffusion models (LDM), introduced by Rombach \etal \cite{rombach2022high}, proposed transferring both diffusion and sampling processes from image space to the latent space of an autoencoder. Given an RGB source image $\mathbf{x}_{0}\in\mathbb{R}^{H\times W\times 3}$, an encoder $\mathcal{E}$ encodes it into a latent representation $\mathbf{z}_{0}=\mathcal{E}(\mathbf{x})\in\mathbb{R}^{h\times{w}\times{c}}$. A decoder $\mathcal{D}$ is jointly-trained to recover the initial images from their respective latents. The encoder downsamples input images by a factor $f=H/h=W/w=2^{m},m\in\mathbb{N}^{+}$.  Subsequently, diffusion can be applied on the produced latents via either DDPM of DDIM, resulting in noisy latents $\mathbf{z}_{t}$, while sampling can still be described with either \cref{eq:ddpm_sampling} or \cref{eq:ddim_sampling}, after substituting images $\mathbf{x}_{(\cdot)}$ with their respective latents $\mathbf{z}_{(\cdot)}$. Moreover, the noise approximation network can be conditioned on additional user inputs $y$, \ie $\bm{\epsilon}_{\bm{\theta}}(\mathbf{z}_{t},t,\bm{\tau}_{\bm{\theta}}(y))$, where $\bm{\tau}_{\bm{\theta}}$ denotes a domain-specific encoder that projects $y$ to an intermediate representation $\bm{\tau}_{\bm{\theta}}(y)\in\mathbb{R}^{M\times{d_{\tau}}}$.

Emotion manipulation with diffusion models currently constitutes an unexplored research area, with few available pieces of related literature. Kim \etal \cite{kim2022diffface} introduced DiffFace, a face swapping framework capable of transferring target attributes such as gaze, structure and pose, while preserving source identities. Wang \etal \cite{wang2022hs} tackled the head-swapping task with HS-Diffusion. More recently, Li \etal \cite{li2023styo} proposed StyO for one-shot face stylization. Text-guided image manipulation and editing \cite{kim2022diffusionclip, chandramouli2022ldedit, parmar2023zero} has also demonstrated competitive results in zero-shot settings.
\vspace{-0.3cm}
\section{Method}
We trained a VQGAN \cite{esser2021taming} as the first stage of our LDM. The LDM is trained using classifier-free guidance \cite{ho2021classifierfree}, dropout probability $p_{\textrm{uncond}}=0.2$ and $T_{\textrm{DDPM}}=1000$ steps. For more details, we refer to the Suppl. Mat.
\subsection{Baseline}
The baseline methodology towards image-based emotion manipulation follows the paradigm of LDEdit \cite{chandramouli2022ldedit}, adapted specifically for the task of emotion translation. In our case, the transformer tokenizer is replaced with a class-embedder that maps each emotion label to a learnable embedding vector $\bm{\tau}_{\bm{\theta}}(y)\in\mathbb{R}^{1\times 512}$. More specifically, the backbone of our approach resides in the near cycle-consistency induced by the deterministic forward and backward DDIM processes $(\eta=0)$. This is further demonstrated by the following analysis. The deterministic DDIM sampling process described in section \cref{eq:ddim_sampling}, in the context of LDMs, can be rewritten in the following form:
\begin{equation}\label{eq:ode}
    \resizebox{.9\hsize}{!}{$
    \sqrt{\frac{1}{\bar{a}_{t-1}}}\mathbf{z}_{t-1}-\sqrt{\frac{1}{\bar{a}_{t}}}\mathbf{z}_{t}=\Bigg(\sqrt{\frac{1}{\bar{a}_{t-1}}-1}-\sqrt{\frac{1}{\bar{a}_{t}}-1}\Bigg)\bm{\epsilon}_{\bm{\theta}}(\mathbf{z}_{t},t)$}
\end{equation}
By setting $\mathbf{y}_{t}:=\sqrt{1/\bar{a}_{t}}\mathbf{z}_{t}$ and $p_{t}:=\sqrt{1/\bar{a}_{t}-1}$, \cref{eq:ode} can be written as:
\begin{equation}
    \resizebox{.9\hsize}{!}{$
    \mathbf{y}_{t-1}-\mathbf{y}_{t}=(p_{t-1}-p_{t})\bm{\epsilon}_{\bm{\theta}}(\mathbf{z}_{t},t)\Rightarrow \mathrm{d}\mathbf{y}_{t}=\bm{\epsilon}_{\bm{\theta}}(\mathbf{z}_{t},t)\mathrm{d}p_{t}$}
\end{equation}
The last equation describes an ODE that moves backward in terms of the diffusion time index $t$. Reversal of this ODE leads to the respective forward deterministic DDIM process.
% \begin{equation}
%     \mathbf{y}_{t+1}-\mathbf{y}_{t}=(p_{t+1}-p_{t})\bm{\epsilon}_{\bm{\theta}}(\mathbf{z}_{t},t)
% \end{equation}

Now, given a source image-label pair $(\mathbf{x}_{\textrm{src}},y_{\textrm{src}})$, its translation w.r.t. a target emotion $y_{\textrm{trg}}$ can be achieved through the following pair of equations:
\begin{align}
    \text{\scriptsize$\mathbf{z}_{t+1}$}&\text{\scriptsize$=\sqrt{\bar{a}_{t+1}}\mathbf{f}_{\bm{\theta}}(\mathbf{z}_{t},t,\bm{\tau}_{\bm{\theta}}(y_{\textrm{src}}))+\sqrt{1-\bar{a}_{t+1}}\mathbf{\epsilon}_{\bm{\theta}}
    (\mathbf{z}_{t},t,\bm{\tau}_{\bm{\theta}}(y_{\textrm{src}}))$} \label{eq:fwd} \\
    \text{\scriptsize$\mathbf{\hat{z}}_{t-1}$}&\text{\scriptsize$=\sqrt{\bar{a}_{t-1}}\mathbf{f}_{\bm{\theta}}(\mathbf{\hat{z}}_{t},t,\bm{\tau}_{\bm{\theta}}(y_{\textrm{trg}}))+\sqrt{1-\bar{a}_{t-1}}\mathbf{\epsilon}_{\bm{\theta}}(\mathbf{\hat{z}}_{t},t,\bm{\tau}_{\bm{\theta}}(y_{\textrm{trg}}))$} \label{eq:rev} 
\end{align}
%During the manipulation processes, LDM weights $\bm{\theta}$ remain frozen in accordance with the best model checkpoint, as obtained from the initial training phase. 
Given the input source image $\mathbf{x}_{\textrm{src}}$, the first stage encoder $\mathcal{E}$ transforms it into its corresponding latent representation $\mathbf{z}_{0}$. Then deterministic forward diffusion is performed up to timestep $t_{0}<T_{\textrm{DDPM}}$, conditioned on the source emotion label $y_{\textrm{src}}$, denoted as $\mathbf{z}_{t_{0}}$, according to \cref{eq:fwd}. The reverse process conditioned on the target emotion label $y_{\textrm{trg}}$ initiates from the same noised latent code $\mathbf{z}_{t_{0}}$ with the aim of reconstructing $\hat{\mathbf{z}}_{0}$. Passing the manipulated latent code $\hat{\mathbf{z}}_{0}$ through the decoder $\mathcal{D}$ results in the reconstructed target image $\mathbf{x}_{\textrm{gen}}$. DDIM processes can be significantly sped up by using fewer discretization steps $\{\tau_{s}\}_{s=1}^{T_{\textrm{DDIM}}}$ evenly selected in the range $[1,t_{0}]$ such that $\tau_{1}=1$ and $\tau_{T_{\textrm{DDIM}}}=t_{0}$. The effects of varying $t_{0}$ are illustrated in \cref{fig:cover}.
%Obviously, the lower the number of intermediate steps $T_{\textrm{DDIM}}$, the worse the reconstruction quality of the DDIM process. 
Lowering the number of sampling steps is essential for speeding up experiments and in practice $T_{\textrm{DDIM}}\geq20$ provides satisfactory results in the context of image-to-image translation.

\subsection{CLIP-guided finetuning}
In order to effectively extract knowledge from CLIP \cite{radford2021learning} in the context of image-translation, two different losses have been proposed: a global target loss \cite{patashnik2021styleclip}, and local directional loss \cite{gal2022stylegan}. The global CLIP loss tries to minimize the cosine distance in CLIP space between the generated image and a given target text. On the other hand, the local directional loss is designed to alleviate issues of the latter such as low diversity and susceptibility to adversarial attacks. The local directional CLIP loss induces the direction between the embeddings of the reference and generated images to be aligned with the direction between the embeddings of a pair of reference and target texts in the CLIP space as follows:
\begin{equation}
    \begin{aligned}
        \mathcal{L}_{\textrm{dir}}=1-\textrm{cos}\bigl\langle&\textrm{CLIP}_{\textrm{img}}(\mathbf{x}_{\textrm{gen}})-\textrm{CLIP}_{\textrm{img}}(\mathbf{x}_{\textrm{src}}), \\
        &\textrm{CLIP}_{\textrm{text}}(y_{\textrm{trg}})-\textrm{CLIP}_{\textrm{text}}(y_{\textrm{src}})\bigr\rangle
    \end{aligned}
\end{equation}
We adapt DiffusionCLIP \cite{kim2022diffusionclip} to the context of image-based emotion translation. At the reverse path, our pre-trained LDM is being finetuned to generate samples driven by target texts $y_{\textrm{trg}}$. Apart from the CLIP component, the overall finetuning objective is comprised of an identity preservation loss, implemented as the cosine distance in the embedding space of a pre-trained IR-SE50 \cite{he2016deep, hu2018squeeze} model coupled with ArcFace \cite{deng2019arcface}, along with a $\ell_{2}$ loss in pixel space:
\begin{equation}
    \mathcal{L}=\lambda_{\textrm{dir}}\mathcal{L}_{\textrm{dir}}+\lambda_{\textrm{id}}\mathcal{L}_{\textrm{id}}(\mathbf{x}_{\textrm{gen}},\mathbf{x}_{\textrm{src}})+\lambda_{\ell_2}\lVert\mathbf{x}_{\textrm{gen}}-\mathbf{x}_{\textrm{src}}\rVert_{2}^{2}
\end{equation}

Our implementation differs from the original one in the following two ways: i) Instead of DDPM, we use LDM as our backbone. ii) The latter did not take advantage of source class-label information when applying the directional CLIP loss, as the majority of experiments were performed in unconditional settings. 
%Moreover, the text editing direction in the original implementation was static, \ie the same for all input images, regardless of their context. 
Moreover, taking advantage of the available class labels, we allowed images to be manipulated in effectively 36 different directions, that is \textrm{`neutral'}$\rightarrow\{$\textrm{`happy', `sad', `surprised', `scared', `disgusted', `angry'}$\}$ and $y_{\textrm{src}}\rightarrow{y_{\textrm{trg}}}$, with $y_{\textrm{trg}}\in\{$\textrm{`happy', `sad', `surprised', `scared', `disgusted', `angry'}$\}$ and $y_{\textrm{src}}\neq y_{\textrm{trg}}$ (contempt was mapped to neutral). As our LDM was trained using \textit{c.f.g.}, we are able to couple CLIP guidance with different values of unconditional guidance scale $\gamma$.
\vspace{-0.2cm}
\section{Experiments}
\vspace{-0.1cm}
\subsection{Dataset \& Evaluation Metrics}
We train and evaluate our models on AffectNet \cite{mollahosseini2017affectnet}. AffectNet is by far the largest database of facial expressions that provides both categorical and VA
annotations, in-the-wild. 
% Out of the 440K manually annotated images, 291,651 images are labeled with 8 basic discrete emotions, \ie neutral, happy, sad, surprised, scared, disgusted, angry and contemptuous. Out of those, 287,651 images belong to the training set and 4,000 images (500 for emotion label) belong to the validation set. 
In the context of image-to-image translation, PSNR, SSIM \cite{wang2004image} and LPIPS \cite{zhang2018unreasonable} constitute the most commonly used evaluation metrics relative to image quality. We also employ HSEmotion \cite{savchenko2021facial, savchenko2022classifying, savchenko2022video}, a SOTA lightweight emotion recognition framework for evaluating the emotion transfer capabilities of our models in terms of classification accuracy. Lastly, we evaluate models based on subject identity preservation using feature cosine similarity (CSIM) in the embedding space of a CosFace \cite{wang2018cosface} model, pre-trained on the Glint360K \cite{an2021partial} dataset.

\begin{table*}[t!]
\centering
\resizebox{15cm}{!}{
\begin{tabular}{c|ccccc|ccccc|ccccc}
\multirow{2}{*}{\diagbox{Method}{$y_{\textrm{target}}$}} & \multicolumn{5}{c|}{Happy}                                                         & \multicolumn{5}{c|}{Sad}                                                           & \multicolumn{5}{c}{Surprised}                                                      \\ \cline{2-16} 
                        & Accuracy$\uparrow$       & PSNR$\uparrow$           & SSIM$\uparrow$           & LPIPS$\downarrow$          & CSIM$\uparrow$           & Accuracy$\uparrow$       & PSNR$\uparrow$           & SSIM$\uparrow$           & LPIPS$\downarrow$          & CSIM$\uparrow$           & Accuracy$\uparrow$       & PSNR$\uparrow$           & SSIM$\uparrow$           & LPIPS$\downarrow$          & CSIM$\uparrow$           \\ \hline
Groundtruth \cite{mollahosseini2017affectnet}             & 0.758          & --              & --              & --              & --             & 0.638          & --              & --              & --              & --              & 0.606          & --              & --             & --              & --              \\
GANimation \cite{pumarola2018ganimation}             & 0.645          & 24.07          & 0.816          & 0.099          & 0.547          & 0.212          & 24.52          & 0.830          & 0.097          & 0.582          & 0.360          & 23.82          & 0.817          & 0.101          & 0.559          \\
StarGAN v2 \cite{choi2020stargan}             & \textbf{0.958} & 17.46          & 0.659          & 0.165          & 0.441          & 0.569          & 18.30          & 0.712          & 0.149          & 0.593          & 0.761          & 17.99          & 0.678          & 0.160          & 0.503          \\
GANmut \cite{d2021ganmut}                 & 0.879          & 21.42          & 0.809          & 0.106          & 0.663          & 0.888          & 23.85          & \textbf{0.857} & 0.094          & 0.755          & 0.829          & 22.43          & 0.810          & 0.112          & 0.675          \\
GGANmut \cite{d2021ganmut}                & 0.934          & 21.91          & 0.819          & 0.103          & 0.717          & \textbf{0.986} & 22.13          & 0.802          & 0.115          & 0.653          & \textbf{0.970} & 22.37          & 0.777          & 0.121          & 0.610          \\
Ours                    & 0.872          & \textbf{25.80} & \textbf{0.841} & \textbf{0.090} & 0.743          & 0.774          & \textbf{25.71} & 0.837          & 0.095          & 0.778          & 0.658          & \textbf{25.54} & \textbf{0.838} & \textbf{0.094} & 0.716          \\
Ours w/ CGF             & 0.883          & 24.30          & 0.813          & 0.098          & \textbf{0.744} & 0.875          & 24.72          & 0.822          & \textbf{0.093} & \textbf{0.794} & 0.752          & 23.42          & 0.797          & 0.113          & \textbf{0.721} \\ \hline
                        & \multicolumn{5}{c|}{Scared}                                                        & \multicolumn{5}{c|}{Disgusted}                                                     & \multicolumn{5}{c}{Angry}                                                          \\ \hline
Groundtruth \cite{mollahosseini2017affectnet}            & 0.666          & --              & --              & --              & --              & 0.646          & --              & --              & --              & --              & 0.514          & --              & --              & --              & --              \\
GANimation \cite{pumarola2018ganimation}             & 0.314          & 23.68          & 0.814          & 0.105          & 0.556          & 0.271          & 24.13          & 0.819          & 0.103          & 0.549          & 0.287          & 24.80          & 0.833          & 0.096          & 0.580          \\
StarGAN v2 \cite{choi2020stargan}             & 0.860          & 17.76          & 0.676          & 0.162          & 0.507          & 0.879          & 18.10          & 0.691          & 0.154          & 0.528          & 0.666          & 18.14          & 0.689          & 0.154          & 0.509          \\
GANmut \cite{d2021ganmut}                 & 0.932          & 23.39          & \textbf{0.841} & 0.102          & 0.721          & 0.877          & 22.64          & 0.815          & 0.113          & 0.663          & 0.856          & 21.57          & 0.813          & 0.106          & 0.678          \\
GGANmut \cite{d2021ganmut}                & \textbf{0.967} & 20.13          & 0.763          & 0.135          & 0.556          & \textbf{0.987} & 21.68          & 0.772          & 0.128          & 0.562          & \textbf{0.969} & 21.77          & 0.767          & 0.133          & 0.590          \\
Ours                    & 0.764          & \textbf{24.89} & 0.824          & 0.103          & \textbf{0.770} & 0.676          & \textbf{25.50} & \textbf{0.835} & 0.100          & 0.707          & 0.710          & \textbf{25.66} & \textbf{0.840} & \textbf{0.094} & 0.714          \\
Ours w/ CGF             & 0.764          & 24.83          & 0.826          & \textbf{0.096} & 0.764          & 0.450          & 24.87          & 0.822          & \textbf{0.089} & \textbf{0.714} & 0.886          & 24.18          & 0.796          & 0.106          & \textbf{0.735} \\ \hline
\end{tabular}}
\caption{Quantitative comparison among our method and GAN-based implementations for image-based emotional manipulation on the AffectNet validation set. Aggregated results can be found in the Suppl. Mat.}
\label{tab:comparisons_main}
\vspace{-0.25cm}
\end{table*}

% \begin{figure}[t!]
%     \centering
%     \begin{minipage}{0.5\textwidth}
%         \centering
%         \includegraphics[width=0.62\textwidth]{img/comparison.jpg} 
%     \end{minipage}%
%     \caption{Qualitative comparison among our method and GAN-based implementations for image-based emotional manipulation on curated examples from the AffectNet validation set.}
%     \label{fig:comparison}
% \end{figure}

\begin{figure*}[t!]
    \centering
    \begin{minipage}{0.8\textwidth}
        \centering
        \includegraphics[width=1\textwidth]{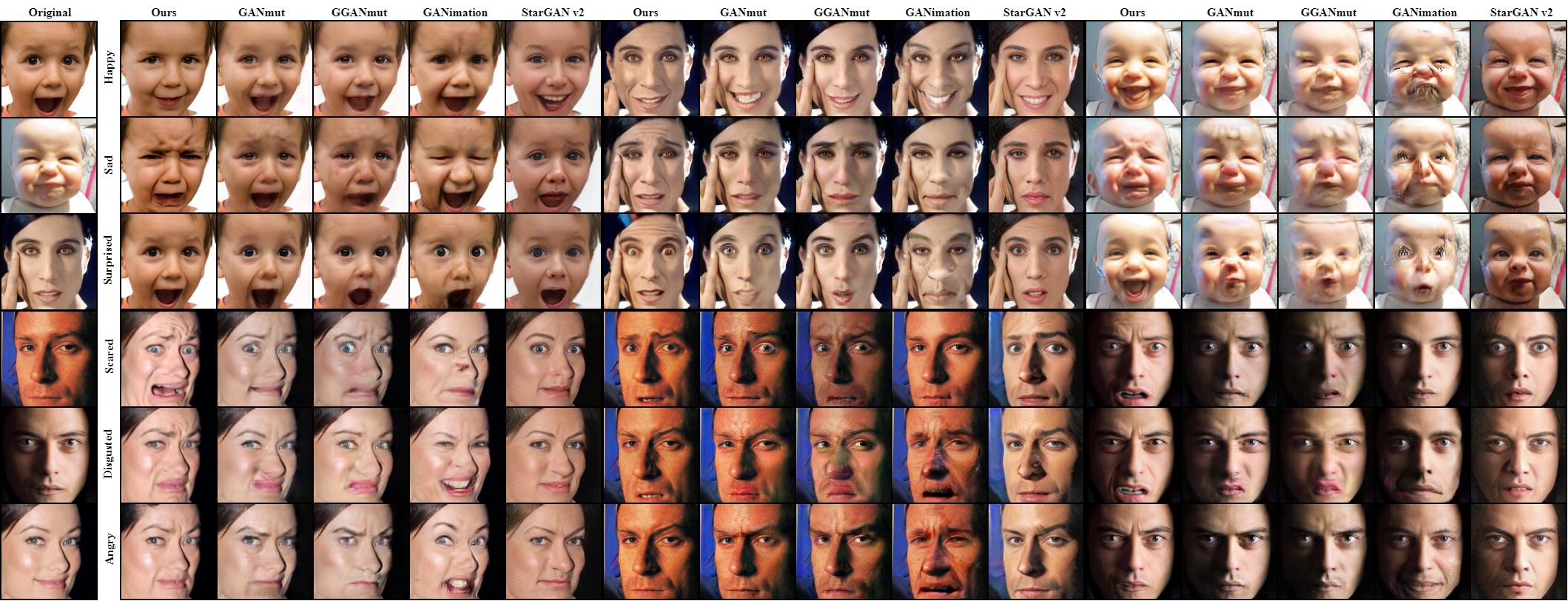} 
    \end{minipage}%
    \caption{Qualitative comparison among our method and GAN-based implementations for image-based emotional manipulation on curated examples from the AffectNet validation set.}
    \label{fig:comparison_horizontal}
\vspace{-0.5cm}
\end{figure*}
\vspace{-0.08cm}
\subsection{Results}
As at the time of writing, no applicable diffusion-based model with a publicly available codebase exists, we shift our attention towards well-grounded GAN-based implementations, with which we will compare our findings, in both a quantitative and qualitative level. These models are: GANimation \cite{pumarola2018ganimation}, StarGAN v2 \cite{choi2020stargan} and GANmut \cite{d2021ganmut} (linear and Gaussian models, denoted as GANmut and GGANmut, respectively). 
%Details about used code and training specifications are included as supplementary material. 
For details about used code and training specifications, please refer to our Suppl. Mat. 
\vspace{-0.5cm}
\paragraph{Objective Results}
In \cref{tab:comparisons_main} we provide quantitative comparisons among the three aforementioned GAN-based implementations and our diffusion-based ones, with and without CLIP-guided finetuning.
%Boldface indicates the best score relative to each table column. 
%Making comparisons between GAN and diffusion-based models is far from straightforward and we need to take into consideration the underlying trade-off between emotion classification accuracy and quality/identity preservation with regard to source images. 
Across all emotions, our models consistently (with few exceptions) surpass all GAN-based counterparts in terms of image quality and subject identity preservation. Our method also achieves higher emotion classification than GANimation and comparable with StarGAN v2 but is outperformed by GGANmut. This can be partly explained by the fact the the latter fits multiple Gaussians with the aim of acquiring a more accurate and rich representation of the conditional emotional latent space, compared to discrete emotion labels.
\vspace{-0.5cm}
\paragraph{User Studies}
We conducted two user studies to evaluate the realism and emotion accuracy of our approach, as assessed by human users, the results of which can be found in \cref{tab:subj_comparison}. In the first study, we involved 24 participants, who were presented with 28 pairwise comparisons between all methods (including original images) and were asked to choose the most realistic one. As demonstrated by the results, our method's manipulated images were perceived \textit{significantly} ($p<0.01$ with binomial test) more realistic compared to the other methods. However, as anticipated, they were not as realistic as the actual images.
 
In the second study, 27 participants were shown 30 images from each method and asked to identify the displayed emotion from a list of six emotions. The bottom half of \cref{tab:subj_comparison} presents the corresponding accuracy results, which are closely aligned with the previous objective results. Our approach achieves performance on-par with StarGAN v2 and surpasses GANimation, although GANmut achieves the highest accuracy. Notably, the accuracy in the original images of AffectNet is similar to that of StarGAN v2 and our method. This observation can potentially be attributed to GANmut generating more "exaggerated" emotions that deviate from the emotion distribution in AffectNet, sacrificing realism in the process. Finally, a qualitative comparison is presented in \cref{fig:comparison_horizontal}, where the superiority of our models in terms of image quality and realism becomes more evident. For further experimental results and details, we kindly direct the reader to our Suppl. Mat.

% \begin{table}[t!]
% \centering
% \resizebox{7cm}{!}{
% \begin{tabular}{ccccccccc}
% \multicolumn{1}{c|}{}         & \multicolumn{8}{c}{Realism}                                                                                                                                         \\ \cline{2-9} 
% \multicolumn{1}{c|}{Ours vs.} & \multicolumn{2}{c|}{GANimation}           & \multicolumn{2}{c|}{GANmut}               & \multicolumn{2}{c|}{StarGAN v2}           & \multicolumn{2}{c}{Groundtruth} \\ \hline
% \multicolumn{1}{c|}{}         & \textbf{0.69} & \multicolumn{1}{c|}{0.31} & \textbf{0.68} & \multicolumn{1}{c|}{0.32} & \textbf{0.61} & \multicolumn{1}{c|}{0.39} & 0.28       & \textbf{0.72}      \\ \hline
%                               & \multicolumn{8}{c}{Emotion Recognition}                                                                                                                             \\ \hline
% \multicolumn{1}{c|}{0.57}     & \multicolumn{2}{c|}{0.38}                 & \multicolumn{2}{c|}{\textbf{0.73}}        & \multicolumn{2}{c|}{0.57}                 & \multicolumn{2}{c}{0.59}        \\ \hline
% \end{tabular}
% }
% \caption{Head-to-head subjective study regarding realism (top half) and emotion translation accuracy (bottom half).}
% \label{tab:subj_comparison}
% \vspace{-0.6cm}
% \end{table}

\begin{table}[t!]
\centering
\setlength{\tabcolsep}{4pt}
\resizebox{\linewidth}{!}{
\begin{tabular}{ccccccccc}
\multicolumn{1}{c|}{}         & \multicolumn{8}{c}{Realism}                                                                                                                                         \\ \cline{2-9} 
\multicolumn{1}{c|}{Ours vs.} & \multicolumn{2}{c|}{GANimation}           & \multicolumn{2}{c|}{GANmut}               & \multicolumn{2}{c|}{StarGAN v2}           & \multicolumn{2}{c}{Groundtruth} \\ \hline
\multicolumn{1}{c|}{}         & \textbf{0.69 (116)} & \multicolumn{1}{c|}{0.31 (52)} & \textbf{0.68 (114)} & \multicolumn{1}{c|}{0.32 (54)} & \textbf{0.61 (103)} & \multicolumn{1}{c|}{0.39 (65)} & 0.28 (46)       & \textbf{0.72 (122)}      \\ \hline
                              & \multicolumn{8}{c}{Emotion Recognition}                                                                                                                             \\ \hline
\multicolumn{1}{c|}{0.57 (92/162)}     & \multicolumn{2}{c|}{0.38 (61/162)}                 & \multicolumn{2}{c|}{\textbf{0.73 (118/162)}}        & \multicolumn{2}{c|}{0.57 (92/162)}                 & \multicolumn{2}{c}{0.59 (96/162)}        \\ \hline
\end{tabular}
}
\caption{Head-to-head subjective study regarding realism (top half) and emotion translation accuracy (bottom half). Numbers in parentheses denote absolute numbers.}
\label{tab:subj_comparison}
\vspace{-0.6cm}
\end{table}

\vspace{-0.2cm}
\section{Conclusion}
In this paper, we presented our method for photorealistic and identity preserving emotion manipulation of images with the use of diffusion models. Our extensive objective and subjective evaluations on AffectNet showcase the superiority of method in terms of quality, while achieving competitive results regarding emotion translation accuracy against both conventional and dedicated GAN-based methodologies. Hopefully, our work can serve as a breeding ground for further research in the field of diffusion-based face reenactment. 

%%%%%%%%% REFERENCES

{\small
\bibliographystyle{ieee_fullname}
\bibliography{egbib}
}

\setcounter{section}{0}
\renewcommand{\thesection}{\Alph{section}}
%%%%%%%%%% Merge with supplemental materials %%%%%%%%%%
\clearpage
\twocolumn[
  \begin{center}
    {\Large \bfseries Photorealistic and Identity-Preserving Image-Based Emotion Manipulation with Latent Diffusion Models -- Supplementary Material}
  \end{center}
  \vspace{1ex}
]
%%%%%%%%%% Merge with supplemental materials %%%%%%%%%%

% \and
\maketitle

\section{Model Specifications}
\subsection{Face Cropping \& Alignment}
The creators of the AffectNet dataset have provided precomputed face bounding boxes as well as facial landmark keypoint coordinates per image. However, we rather opted for manually locating both the bounding boxes as well as the facial landmarks from scratch, using FAN \cite{bulat2017far}. Before feeding the original images to FAN, we first resize them to $256^{2}$ pixels in order to avoid running out of GPU memory. FAN locates the 2D coordinates for 68 facial landmarks. Based on those landmarks, bounding boxes of size $224^{2}$ are center-cropped and then aligned by estimating the affine transform required to warp the set of detected facial keypoints to their canonical coordinates. Any empty image regions resulting from rotating the original image are dealt with by reflecting the image near the edges. 

\subsection{First Stage}
The used VQGAN \cite{esser2021taming} follows the U-Net \cite{ronneberger2015u} architecture consisting of a series of downsampling and upsampling residual blocks. The number of residual blocks per spatial resolution level was set equal to two. The number of basic channels was set equal to 128. The given input images were first resized to a resolution of $128^{2}$ pixels. Thus, given input images $\mathbf{x}\in\mathbb{R}^{128\times128\times3}$, the encoder compresses the latter by a factor $f=4$, resulting in latents $\hat{\mathbf{z}}\in\mathbb{R}^{32\times32\times3}$. Each latent undergoes element-wise quantization resulting in a collection of $32^{2}$ codebook entries $\bm{z}_{ij}\in\mathbb{R}^{3}, i,j\in[32]$. The cardinality of the entire codebook is set equal to $|\mathcal{Z}|=16384$. Moreover, the channel multipliers for the 2 intermediate spatial resolutions is set equal to 2 and 4 respectively. The last spatial resolution level also includes a spatial attention block. The VQGAN was trained for 6 epochs with the Adam \cite{kingma2014adam} optimizer, an initial learning rate of $3.6\times10^{-5}$ and a batch size of 8, reaching a reconstruction loss equal to 0.138 (sum of $\ell_{1}$ and LPIPS distance) on the AffectNet \cite{mollahosseini2017affectnet} validation set. The decoder's structure is exactly the same as the encoder's, but with the order of the blocks being reversed, resulting in approximately 61M trainable parameters.

\subsection{Denoising Network}
The underlying LDM U-Net architecture features two residual blocks per spatial resolution level, with the number of basic channels being equal to 160. Input latents are further downsampled by a factor of $f=4$. Attention blocks are placed at every spatial resolution level, i.e. $32^{2},16^{2},8^{2}$ and the channel multipliers are set equal to 1, 2 and 4 respectively. Each attention block is implemented according to original cross-attention mechanism described in \cite{rombach2022high}. The class label encoder is implemented as a single learnable embedding layer with a dimensionality of 512, mapping class labels $y$ to $\bm{\tau}_{\bm{\theta}}(y)\in\mathbb{R}^{1\times512}$. All the above led to a total number of 156M trainable parameters.

In order to better control the intensity of the depicted emotions during the manipulation phase, we were obliged to train the LDM using \textit{classifier-free guidance} \cite{ho2021classifierfree}. During training and with a probability $p_{\textrm{uncond}}=0.2$, the input class labels are randomly dropped and replaced with a $\varnothing$ (null) label, effectively raising the total number of class labels to nine. Then sampling is performed using the following linear combination of the conditional
and unconditional score estimates:
\begin{equation}
    \resizebox{.88\hsize}{!}{$
    \tilde{\bm{\epsilon}}_{\bm{\theta}}(\mathbf{z}_{t},\bm{\tau}_{\bm{\theta}}(y))=(1-\gamma)\bm{\epsilon}_{\bm{\theta}}(\mathbf{z}_{t},\bm{\tau}_{\bm{\theta}}(\varnothing))+\gamma\bm{\epsilon}_{\bm{\theta}}(\mathbf{z}_{t},\bm{\tau}_{\bm{\theta}}(y))$}
\end{equation}
The LDM was trained for around 30 epochs with the AdamW \cite{loshchilov2018decoupled} optimizer, an initial learning rate of $2.4\times10^{-5}$ and a batch size of 24. We used a linear noise schedule and $T_{\textrm{DDPM}}=1000$ diffusion steps.
 
\section{Manipulation Hyperparameters}
Important hyperparameters of both the baseline and tuned manipulation processes are the stochasticity parameter $\eta$, the editing strength $t_{0}$, the number of steps $T_{\textrm{DDIM}}$ and the unconditional guidance scale $\gamma$. We begin by setting $\eta=0$ for all of the upcoming experiments. In general, values $\eta>0$ can produce diverse outputs, reducing the consistency with the original image. In the context of facial expression manipulation, identity and facial attribute preservation of the depicted subjects is of primary importance and as the current baseline methodology does not include any other form of supervision upon the latter, we find ourselves constrained to use $\eta=0$. Moreover, we chose to name $t_{0}$ as the editing strength hyperparameter due to the observation that the higher its value, the more prominent the manipulation effect becomes, compared to the original image. As we chose to train the LDM using \textit{c.f.g.}, we have one extra degree of freedom regarding the intensity of the produced manipulation in the form of the unconditional guidance scale $\gamma$.

\begin{figure}[t!]
    \centering
    \begin{minipage}{0.35\textwidth}
        \centering
        \includegraphics[width=1.0\textwidth]{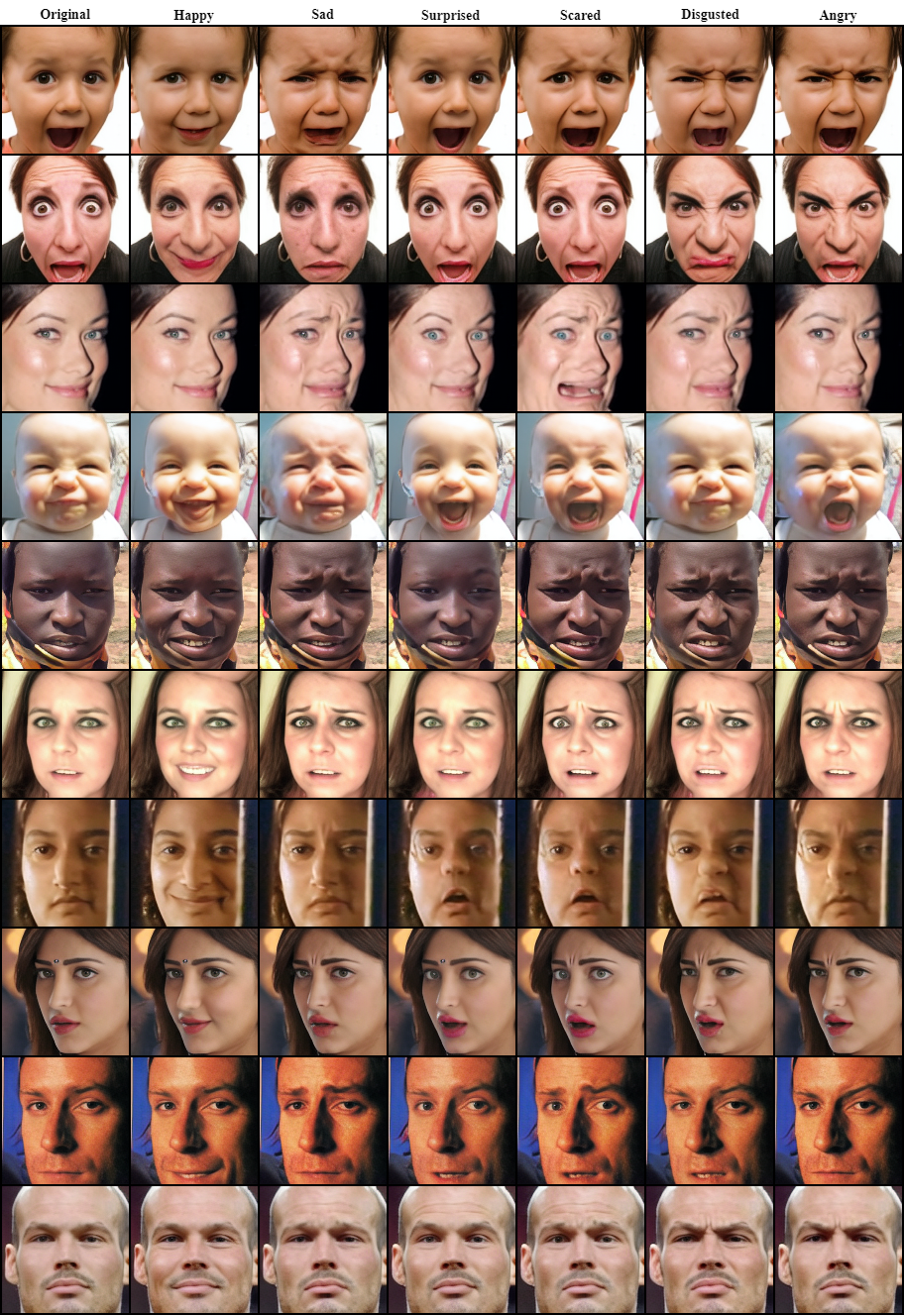} 
    \end{minipage}%
    \caption{Curated examples of latent emotion manipulation on the AffectNet validation set, using $T_{\textrm{DDIM}}=40$ DDIM steps, $\eta=0$, editing strength $t_{0}=500$ and \textit{c.f.g} scale $\gamma=3.0$.}
    \label{fig:baseline_var_sample}
\end{figure}

\begin{figure}
\centering
\begin{subfigure}[b]{0.35\textwidth}
   \includegraphics[width=1\linewidth]{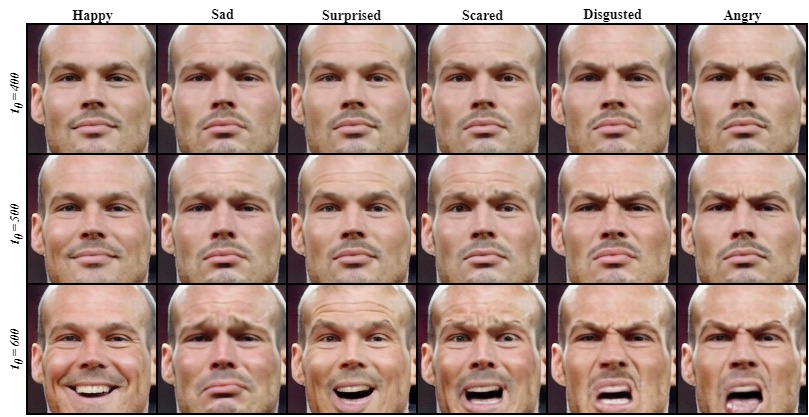}
\end{subfigure}

\begin{subfigure}[b]{0.35\textwidth}
   \includegraphics[width=1\linewidth]{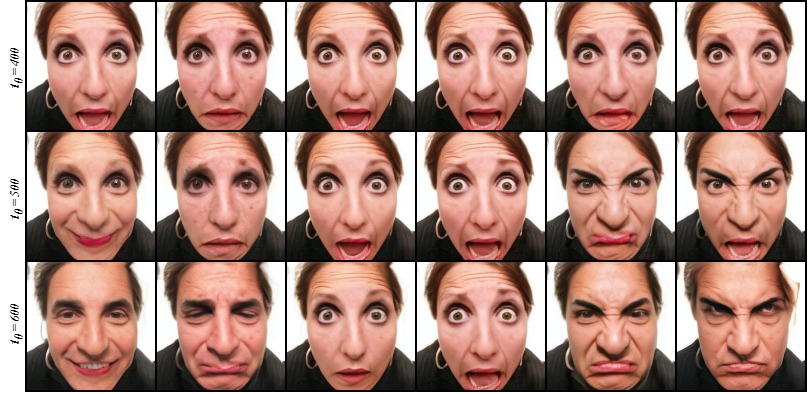}
\end{subfigure}

\begin{subfigure}[b]{0.35\textwidth}
   \includegraphics[width=1\linewidth]{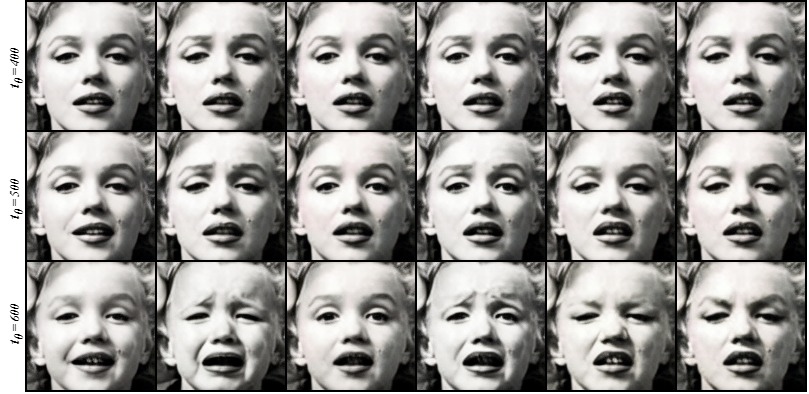}
\end{subfigure}
\caption{Curated examples of LDM-based emotion manipulation on the {AffectNet} validation set, using $T_{\textrm{DDIM}}=40$ steps, $\eta=0$, \textit{c.f.g} scale $\gamma=3.0$ and variable editing strength $t_{0}\in\{400,500,600\}$.}
\label{fig:baseline_var_t0}
\end{figure}

\begin{figure}
\centering
\begin{subfigure}[b]{0.32\textwidth}
   \includegraphics[width=1\linewidth]{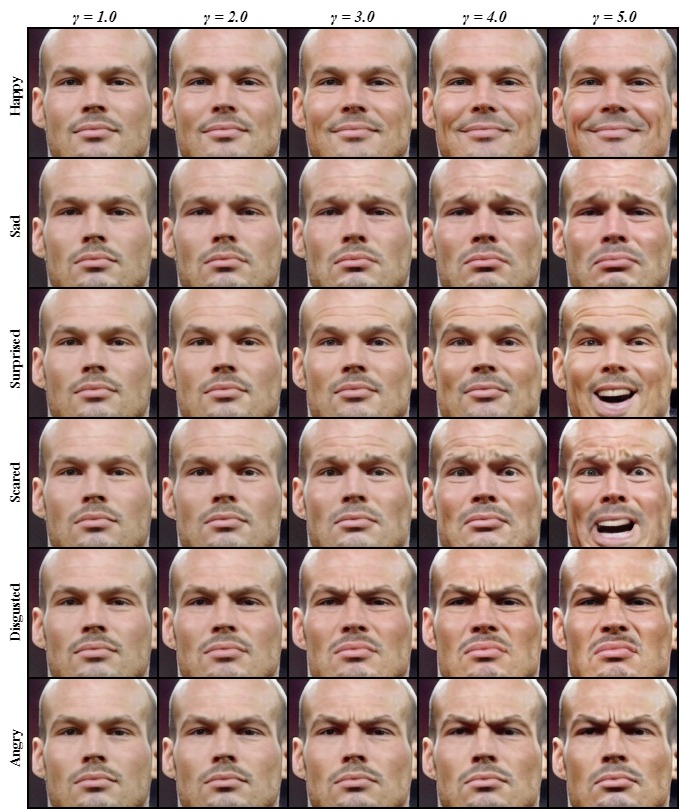}
\end{subfigure}

\begin{subfigure}[b]{0.32\textwidth}
   \includegraphics[width=1\linewidth]{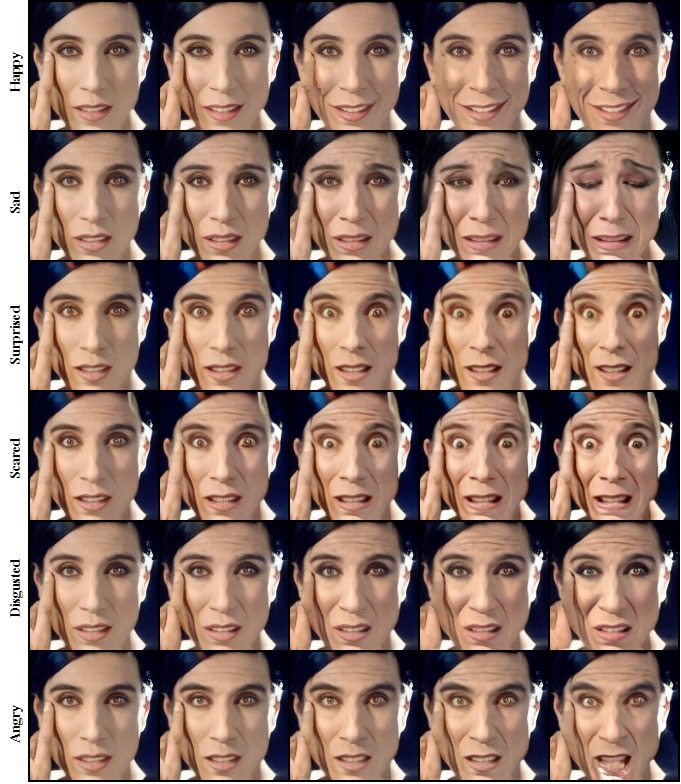}
\end{subfigure}

\begin{subfigure}[b]{0.32\textwidth}
   \includegraphics[width=1\linewidth]{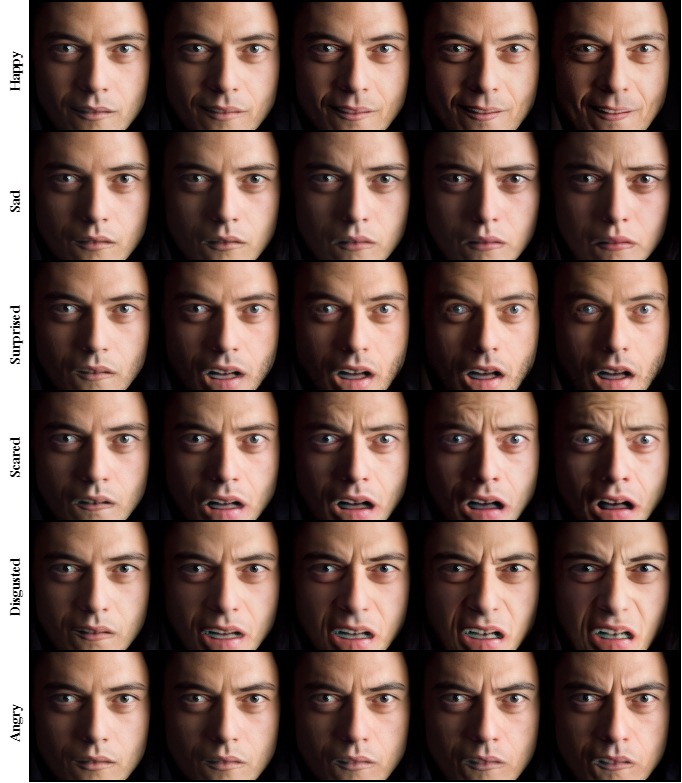}
\end{subfigure}
\caption{Curated examples of LDM-based emotion manipulation on the AffectNet validation set, using $T_{\textrm{DDIM}}=40$ steps, $\eta=0$, editing strength $t_{0}=500$ and variable \textit{c.f.g.} scale $\gamma\in\{1.0,2.0,3.0,4.0,5.0\}$.}
\label{fig:baseline_var_scale}
\end{figure}

\cref{fig:baseline_var_sample} illustrates curated examples of emotion manipulation on images from the AffectNet validation set, using $T_{\textrm{DDIM}}=20$, $\gamma=3.0$ and $t_{0}=500$. The leftmost image in each column is an original sample drawn from the dataset, while the other six columns include the manipulation results for each one of the 6 basic expressions of \textit{happiness}, \textit{sadness}, \textit{surprise}, \textit{fear}, \textit{disgust} and \textit{anger}. We can immediately notice that the results in the `happy', `sad' and `angry' columns are overall satisfactory, the emotions of surprise, fear and disgust are more difficult to convey. This can be partly justified by the fact that the aforementioned emotions are underrepresented in the available AffectNet training set. \cref{fig:baseline_var_t0} illustrates the effect of varying the editing strength $t_{0}\in\{400,500,600\}$ in the case of all six different target emotions, while $\gamma$ and $T_{\textrm{DDIM}}$ are fixed. One can easily notice that $t_{0}=400$ is not sufficient for producing emotionally distinguishable results. On the other end, a value of $t_{0}=600$ produces better manipulation results but at the cost of subject identity distortion. Furthermore, \cref{fig:baseline_var_scale} illustrates the effect of varying the unconditional guidance scale $\gamma\in\{1.0,2.0,3.0,4.0,5.0\}$ in the case of all six different target emotions, while $t_{0}$ and $T_{\textrm{DDIM}}$ are fixed. It is evident that in the specific setting of $t_{0}=500$, lower values of $\gamma$ have little to no effect in changing the emotion of a given subject. Better results can be obtained with $\gamma\in\{4.0, 5.0\}$ with minimal distortion of peripheral facial features and subject identity.

\subsection{Ablation Study}

Detailed quantitative results are presented in \cref{tab:baseline1} and \cref{tab:baseline2}. The quality of the produced images and consistency with the original ones is measured, as usual, in terms of PSNR, SSIM and LPIPS\footnote{PSNR, SSIM and LPIPS are calculated using \href{https://github.com/photosynthesis-team/piq}{\texttt{piq}}.}. We also employ {HSEmotion}\footnote{Ranked \#1 in 8-emotion classification accuracy on {AffectNet}, according to \url{https://paperswithcode.com/sota/facial-expression-recognition-on-affectnet}.} \cite{savchenko2021facial, savchenko2022video, savchenko2022classifying}, a state-of-the-art lightweight emotion recognition framework for evaluating the emotion transfer capabilities of our baseline methodology. In \cref{tab:baseline1}, we notice that the most difficult expressions, as far as emotion transfer is concerned, are (according to {HSEmotion}) surprise and disgust, while the easiest one is happiness. More specifically, a maximum emotion accuracy of 0.961 is achieved in terms of happiness, while 0.731 and 0.734 for surprise and disgust, respectively. As expected, the highest recognition accuracy for all six emotions is achieved using $\gamma=5.0$ and $t_{0}=600$. This setting also results in the lowest image quality, that regardless of the target emotion, revolves around 22.00 dB PSNR, 0.730 SSIM and 0.150 LPIPS. \cref{tab:baseline2}, quantifies the effect of varying the number of steps $T_{\textrm{DDIM}}$ in terms of the three aforementioned image quality metrics and recognition accuracy. For $t_{0}=400$ and regardless of the target emotion, increasing $T_{\textrm{DDIM}}$ leads to either no change or a small drop in recognition performance. This is not the case for higher $t_{0}\in\{500,600\}$. PSNR always increases in accordance with the increase in sampling steps. SSIM behaves differently, as it increases along with $T_{\textrm{DDIM}}$ when $t_{0}\in\{400,500\}$ and decreases when $t_{0}=600$. A similar counter-intuitive behavior is noticed in the case of LPIPS which gradually increases, as the number of DDIM steps increases, when $t_{0}\in\{500,600\}$.

\begin{table*}[ht!]
\centering
\resizebox{11.6cm}{!}{
\begin{tabular}{c|c|c|n{3}{2}cc|ccccccc}\hline
\multirow{2}{*}{$y_{\textrm{target}}$} & \multirow{2}{*}{\textit{c.f.g.} Scale ($\gamma$)} & \multirow{2}{*}{Str. ($t_{0}$)} & \multicolumn{3}{c|}{Image Quality} & \multicolumn{7}{c}{Emotion Recognition Accuracy w/ {HSEmotion} \cite{savchenko2021facial, savchenko2022video, savchenko2022classifying}}                    \\ \cline{4-13} 
                                &                            &                           & PSNR$\uparrow$       & SSIM$\uparrow$      & LPIPS$\downarrow$     & Neutral & Happy & Sad   & Surprised & Disgusted & Disgusted & Angry \\ \hline
\multirow{15}{*}{Happiness}     & 1.0                          & \multirow{5}{*}{400}      & 35.249     & 0.966     & 0.013     & 0.24    & \textcolor{YellowOrange}{0.248} & 0.076 & 0.146     & 0.093     & 0.075     & 0.122 \\
                                & 2.0                          &                           & 33.150     & 0.95      & 0.021     & 0.254   & \textcolor{YellowOrange}{0.344} & 0.047 & 0.154     & 0.058     & 0.052     & 0.092 \\
                                & 3.0                          &                           & 31.706     & 0.937     & 0.029     & 0.244   & \textcolor{YellowOrange}{0.447} & 0.028 & 0.148     & 0.035     & 0.032     & 0.066 \\
                                & 4.0                          &                           & 30.585     & 0.924     & 0.037     & 0.216   & \textcolor{YellowOrange}{0.537} & 0.018 & 0.138     & 0.023     & 0.021     & 0.047 \\
                                & 5.0                          &                           & 29.659     & 0.911     & 0.045     & 0.179   & \textcolor{YellowOrange}{0.626} & 0.01  & 0.121     & 0.016     & 0.018     & 0.031 \\ \cline{2-3}
                                & 1.0                          & \multirow{5}{*}{500}      & 32.088     & 0.939     & 0.03      & 0.223   & \textcolor{YellowOrange}{0.364} & 0.057 & 0.138     & 0.068     & 0.06      & 0.09  \\
                                & 2.0                          &                           & 29.525     & 0.908     & 0.049     & 0.191   & \textcolor{YellowOrange}{0.58}  & 0.02  & 0.115     & 0.025     & 0.025     & 0.043 \\
                                & 3.0                          &                           & 27.923     & 0.883     & 0.065     & 0.13    & \textcolor{YellowOrange}{0.728} & 0.01  & 0.082     & 0.012     & 0.018     & 0.019 \\
                                & 4.0                          &                           & 26.749     & 0.861     & 0.078     & 0.084   & \textcolor{YellowOrange}{0.819} & 0.006 & 0.066     & 0.007     & 0.009     & 0.01  \\
                                & 5.0                          &                           & 25.8       & 0.841     & 0.090     & 0.056   & \textcolor{YellowOrange}{0.872} & 0.003 & 0.05      & 0.004     & 0.007     & 0.007 \\ \cline{2-3}
                                & 1.0                          & \multirow{5}{*}{600}      & 28.385     & 0.889     & 0.062     & 0.163   & \textcolor{YellowOrange}{0.566} & 0.035 & 0.103     & 0.036     & 0.044     & 0.052 \\
                                & 2.0                          &                           & 25.677     & 0.839     & 0.093     & 0.079   & \textcolor{YellowOrange}{0.821} & 0.007 & 0.052     & 0.011     & 0.018     & 0.012 \\
                                & 3.0                          &                           & 24.125     & 0.803     & 0.113     & 0.038   & \textcolor{YellowOrange}{0.916} & 0.003 & 0.027     & 0.004     & 0.008     & 0.004 \\
                                & 4.0                          &                           & 22.995     & 0.773     & 0.129     & 0.022   & \textcolor{YellowOrange}{0.945} & 0.002 & 0.02      & 0.002     & 0.005     & 0.003 \\
                                & 5.0                          &                           & 22.044     & 0.747     & 0.143     & 0.014   & \textcolor{YellowOrange}{0.961} & 0.002 & 0.015     & 0.002     & 0.003     & 0.003 \\ \hline
\multirow{15}{*}{Sadness}       & 1.0                          & \multirow{5}{*}{400}      & 35.391     & 0.967     & 0.012     & 0.202   & 0.144 & \textcolor{YellowOrange}{0.174} & 0.107     & 0.13      & 0.099     & 0.146 \\
                                & 2.0                          &                           & 33.241     & 0.952     & 0.020     & 0.188   & 0.092 & \textcolor{YellowOrange}{0.27}  & 0.079     & 0.144     & 0.094     & 0.135 \\
                                & 3.0                          &                           & 31.676     & 0.937     & 0.029     & 0.159   & 0.052 & \textcolor{YellowOrange}{0.393} & 0.052     & 0.138     & 0.083     & 0.123 \\
                                & 4.0                          &                           & 30.409     & 0.921     & 0.039     & 0.129   & 0.03  & \textcolor{YellowOrange}{0.507} & 0.036     & 0.13      & 0.064     & 0.104 \\
                                & 5.0                          &                           & 29.329     & 0.906     & 0.049     & 0.098   & 0.014 & \textcolor{YellowOrange}{0.609} & 0.022     & 0.121     & 0.05      & 0.086 \\ \cline{2-3}
                                & 1.0                          & \multirow{5}{*}{500}      & 32.718     & 0.946     & 0.025     & 0.212   & 0.104 & \textcolor{YellowOrange}{0.23}  & 0.091     & 0.131     & 0.092     & 0.14  \\
                                & 2.0                          &                           & 30.083     & 0.917     & 0.044     & 0.179   & 0.038 & \textcolor{YellowOrange}{0.408} & 0.048     & 0.126     & 0.077     & 0.124 \\
                                & 3.0                          &                           & 28.282     & 0.890     & 0.061     & 0.128   & 0.014 & \textcolor{YellowOrange}{0.573} & 0.027     & 0.106     & 0.052     & 0.1   \\
                                & 4.0                          &                           & 26.887     & 0.864     & 0.078     & 0.084   & 0.009 & \textcolor{YellowOrange}{0.696} & 0.014     & 0.086     & 0.034     & 0.077 \\
                                & 5.0                          &                           & 25.713     & 0.837     & 0.095     & 0.06    & 0.006 & \textcolor{YellowOrange}{0.774} & 0.008     & 0.067     & 0.025     & 0.06  \\ \cline{2-3}
                                & 1.0                          & \multirow{5}{*}{600}      & 29.853     & 0.913     & 0.047     & 0.213   & 0.064 & \textcolor{YellowOrange}{0.296} & 0.075     & 0.126     & 0.082     & 0.144 \\
                                & 2.0                          &                           & 26.946     & 0.867     & 0.078     & 0.142   & 0.014 & \textcolor{YellowOrange}{0.549} & 0.032     & 0.099     & 0.055     & 0.109 \\
                                & 3.0                          &                           & 25.062     & 0.827     & 0.103     & 0.09    & 0.008 & \textcolor{YellowOrange}{0.712} & 0.017     & 0.069     & 0.029     & 0.076 \\
                                & 4.0                          &                           & 23.587     & 0.789     & 0.126     & 0.052   & 0.005 & \textcolor{YellowOrange}{0.806} & 0.008     & 0.051     & 0.02      & 0.058 \\
                                & 5.0                          &                           & 22.350     & 0.752     & 0.146     & 0.036   & 0.005 & \textcolor{YellowOrange}{0.849} & 0.005     & 0.04      & 0.016     & 0.049 \\ \hline
\multirow{15}{*}{Surprise}      & 1.0                          & \multirow{5}{*}{400}      & 35.452     & 0.967     & 0.012     & 0.205   & 0.183 & 0.089 & \textcolor{YellowOrange}{0.184}     & 0.126     & 0.079     & 0.134 \\
                                & 2.0                          &                           & 33.379     & 0.953     & 0.020     & 0.204   & 0.19  & 0.063 & \textcolor{YellowOrange}{0.252}     & 0.126     & 0.053     & 0.112 \\
                                & 3.0                          &                           & 31.873     & 0.940     & 0.028     & 0.191   & 0.18  & 0.048 & \textcolor{YellowOrange}{0.337}     & 0.11      & 0.04      & 0.094 \\
                                & 4.0                          &                           & 30.639     & 0.927     & 0.037     & 0.168   & 0.173 & 0.035 & \textcolor{YellowOrange}{0.418}     & 0.105     & 0.028     & 0.074 \\
                                & 5.0                          &                           & 29.582     & 0.913     & 0.046     & 0.135   & 0.167 & 0.027 & \textcolor{YellowOrange}{0.482}     & 0.106     & 0.024     & 0.059 \\ \cline{2-3}
                                & 1.0                          & \multirow{5}{*}{500}      & 32.580     & 0.945     & 0.026     & 0.209   & 0.165 & 0.072 & \textcolor{YellowOrange}{0.242}     & 0.137     & 0.063     & 0.112 \\
                                & 2.0                          &                           & 29.901     & 0.916     & 0.045     & 0.174   & 0.148 & 0.044 & \textcolor{YellowOrange}{0.398}     & 0.12      & 0.034     & 0.082 \\
                                & 3.0                          &                           & 28.096     & 0.889     & 0.063     & 0.127   & 0.136 & 0.027 & \textcolor{YellowOrange}{0.522}     & 0.115     & 0.024     & 0.048 \\
                                & 4.0                          &                           & 26.700     & 0.863     & 0.079     & 0.091   & 0.125 & 0.017 & \textcolor{YellowOrange}{0.603}     & 0.12      & 0.014     & 0.03  \\
                                & 5.0                          &                           & 25.536     & 0.838     & 0.094     & 0.066   & 0.114 & 0.014 & \textcolor{YellowOrange}{0.658}     & 0.12      & 0.008     & 0.022 \\ \cline{2-3}
                                & 1.0                          & \multirow{5}{*}{600}      & 29.426     & 0.907     & 0.051     & 0.191   & 0.131 & 0.064 & \textcolor{YellowOrange}{0.335}     & 0.143     & 0.049     & 0.088 \\
                                & 2.0                          &                           & 26.435     & 0.856     & 0.084     & 0.191   & 0.131 & 0.064 & \textcolor{YellowOrange}{0.335}     & 0.143     & 0.049     & 0.088 \\
                                & 3.0                          &                           & 24.448     & 0.813     & 0.110     & 0.078   & 0.085 & 0.016 & \textcolor{YellowOrange}{0.662}     & 0.133     & 0.008     & 0.018 \\
                                & 4.0                          &                           & 22.957     & 0.774     & 0.133     & 0.048   & 0.092 & 0.008 & \textcolor{YellowOrange}{0.714}     & 0.122     & 0.006     & 0.012 \\
                                & 5.0                          &                           & 21.726     & 0.738     & 0.152     & 0.036   & 0.1   & 0.006 & \textcolor{YellowOrange}{0.731}     & 0.115     & 0.004     & 0.009 \\ \hline
\multirow{15}{*}{Fear}          & 1.0                          & \multirow{5}{*}{400}      & 35.325     & 0.967     & 0.012     & 0.164   & 0.15  & 0.122 & 0.148     & \textcolor{YellowOrange}{0.174}     & 0.093     & 0.148 \\
                                & 2.0                          &                           & 33.121     & 0.952     & 0.020     & 0.124   & 0.108 & 0.13  & 0.149     & \textcolor{YellowOrange}{0.272}     & 0.088     & 0.13  \\
                                & 3.0                          &                           & 31.471     & 0.937     & 0.030     & 0.087   & 0.071 & 0.124 & 0.14      & \textcolor{YellowOrange}{0.388}     & 0.081     & 0.109 \\
                                & 4.0                          &                           & 30.114     & 0.921     & 0.040     & 0.053   & 0.048 & 0.115 & 0.13      & \textcolor{YellowOrange}{0.498}     & 0.066     & 0.09  \\
                                & 5.0                          &                           & 28.935     & 0.904     & 0.051     & 0.033   & 0.032 & 0.098 & 0.113     & \textcolor{YellowOrange}{0.598}     & 0.057     & 0.068 \\ \cline{2-3}
                                & 1.0                          & \multirow{5}{*}{500}      & 32.566     & 0.946     & 0.025     & 0.151   & 0.113 & 0.118 & 0.166     & \textcolor{YellowOrange}{0.222}     & 0.093     & 0.136 \\
                                & 2.0                          &                           & 29.750     & 0.915     & 0.045     & 0.085   & 0.058 & 0.117 & 0.17      & \textcolor{YellowOrange}{0.396}     & 0.066     & 0.109 \\
                                & 3.0                          &                           & 27.750     & 0.884     & 0.065     & 0.043   & 0.033 & 0.092 & 0.148     & \textcolor{YellowOrange}{0.562}     & 0.05      & 0.072 \\
                                & 4.0                          &                           & 26.185     & 0.854     & 0.084     & 0.02    & 0.022 & 0.072 & 0.112     & \textcolor{YellowOrange}{0.692}     & 0.035     & 0.048 \\
                                & 5.0                          &                           & 24.885     & 0.824     & 0.103     & 0.012   & 0.017 & 0.054 & 0.093     & \textcolor{YellowOrange}{0.764}     & 0.028     & 0.033 \\ \cline{2-3}
                                & 1.0                          & \multirow{5}{*}{600}      & 29.539     & 0.910     & 0.049     & 0.133   & 0.073 & 0.114 & 0.195     & \textcolor{YellowOrange}{0.291}     & 0.073     & 0.122 \\
                                & 2.0                          &                           & 26.260     & 0.855     & 0.085     & 0.052   & 0.024 & 0.089 & 0.18      & \textcolor{YellowOrange}{0.541}     & 0.044     & 0.071 \\
                                & 3.0                          &                           & 24.115     & 0.806     & 0.115     & 0.016   & 0.016 & 0.062 & 0.136     & \textcolor{YellowOrange}{0.708}     & 0.023     & 0.039 \\
                                & 4.0                          &                           & 22.492     & 0.760     & 0.141     & 0.009   & 0.014 & 0.045 & 0.096     & \textcolor{YellowOrange}{0.794}     & 0.017     & 0.024 \\
                                & 5.0                          &                           & 21.159     & 0.717     & 0.164     & 0.008   & 0.012 & 0.038 & 0.081     & \textcolor{YellowOrange}{0.832}     & 0.013     & 0.016 \\ \hline
\multirow{15}{*}{Disgust}       & 1.0                          & \multirow{5}{*}{400}      & 35.378     & 0.967     & 0.012     & 0.177   & 0.151 & 0.111 & 0.122     & 0.122     &\textcolor{YellowOrange}{ 0.144}     & 0.173 \\
                                & 2.0                          &                           & 33.184     & 0.952     & 0.020     & 0.149   & 0.106 & 0.108 & 0.1       & 0.116     & \textcolor{YellowOrange}{0.224}     & 0.197 \\
                                & 3.0                          &                           & 31.545     & 0.937     & 0.030     & 0.115   & 0.073 & 0.098 & 0.074     & 0.111     & \textcolor{YellowOrange}{0.317}     & 0.212 \\
                                & 4.0                          &                           & 30.211     & 0.921     & 0.040     & 0.094   & 0.05  & 0.082 & 0.056     & 0.094     & \textcolor{YellowOrange}{0.413}     & 0.211 \\
                                & 5.0                          &                           & 29.068     & 0.904     & 0.051     & 0.065   & 0.032 & 0.076 & 0.041     & 0.08      & \textcolor{YellowOrange}{0.509}     & 0.196 \\ \cline{2-3}
                                & 1.0                          & \multirow{5}{*}{500}      & 32.759     & 0.947     & 0.024     & 0.17    & 0.122 & 0.112 & 0.104     & 0.11      & \textcolor{YellowOrange}{0.19}      & 0.192 \\
                                & 2.0                          &                           & 30.099     & 0.919     & 0.043     & 0.119   & 0.064 & 0.088 & 0.069     & 0.099     & \textcolor{YellowOrange}{0.335}     & 0.226 \\
                                & 3.0                          &                           & 28.214     & 0.891     & 0.062     & 0.073   & 0.04  & 0.072 & 0.04      & 0.073     & \textcolor{YellowOrange}{0.478}     & 0.224 \\
                                & 4.0                          &                           & 26.745     & 0.863     & 0.081     & 0.044   & 0.029 & 0.057 & 0.027     & 0.055     & \textcolor{YellowOrange}{0.594}     & 0.194 \\
                                & 5.0                          &                           & 25.497     & 0.835     & 0.100     & 0.029   & 0.025 & 0.042 & 0.017     & 0.043     & \textcolor{YellowOrange}{0.676}     & 0.168 \\ \cline{2-3}
                                & 1.0                          & \multirow{5}{*}{600}      & 29.914     & 0.915     & 0.047     & 0.15    & 0.095 & 0.097 & 0.083     & 0.099     & \textcolor{YellowOrange}{0.247}     & 0.229 \\
                                & 2.0                          &                           & 26.948     & 0.869     & 0.078     & 0.082   & 0.042 & 0.069 & 0.043     & 0.069     & \textcolor{YellowOrange}{0.452}     & 0.243 \\
                                & 3.0                          &                           & 24.981     & 0.826     & 0.106     & 0.038   & 0.033 & 0.047 & 0.02      & 0.052     & \textcolor{YellowOrange}{0.593}     & 0.216 \\
                                & 4.0                          &                           & 23.489     & 0.787     & 0.131     & 0.021   & 0.034 & 0.036 & 0.012     & 0.039     & \textcolor{YellowOrange}{0.682}     & 0.176 \\
                                & 5.0                          &                           & 22.237     & 0.750     & 0.154     & 0.015   & 0.036 & 0.028 & 0.009     & 0.031     & \textcolor{YellowOrange}{0.734}     & 0.147 \\ \hline
\multirow{15}{*}{Anger}         & 1.0                          & \multirow{5}{*}{400}      & 35.566     & 0.968     & 0.011     & 0.203   & 0.146 & 0.102 & 0.119     & 0.115     & 0.105     & \textcolor{YellowOrange}{0.21}  \\
                                & 2.0                          &                           & 33.477     & 0.954     & 0.019     & 0.188   & 0.102 & 0.091 & 0.104     & 0.1       & 0.124     & \textcolor{YellowOrange}{0.291} \\
                                & 3.0                          &                           & 31.943     & 0.940     & 0.027     & 0.168   & 0.071 & 0.085 & 0.084     & 0.088     & 0.132     & \textcolor{YellowOrange}{0.372} \\
                                & 4.0                          &                           & 30.698     & 0.926     & 0.036     & 0.146   & 0.044 & 0.076 & 0.072     & 0.073     & 0.141     & \textcolor{YellowOrange}{0.448} \\
                                & 5.0                          &                           & 29.627     & 0.912     & 0.045     & 0.127   & 0.03  & 0.068 & 0.057     & 0.064     & 0.142     & \textcolor{YellowOrange}{0.512} \\ \cline{2-3}
                                & 1.0                          & \multirow{5}{*}{500}      & 32.761     & 0.946     & 0.025     & 0.201   & 0.102 & 0.098 & 0.104     & 0.101     & 0.125     & \textcolor{YellowOrange}{0.269} \\
                                & 2.0                          &                           & 30.115     & 0.918     & 0.043     & 0.168   & 0.041 & 0.077 & 0.076     & 0.071     & 0.13      & \textcolor{YellowOrange}{0.436} \\
                                & 3.0                          &                           & 28.304     & 0.891     & 0.061     & 0.127   & 0.021 & 0.063 & 0.05      & 0.054     & 0.128     & \textcolor{YellowOrange}{0.558} \\
                                & 4.0                          &                           & 26.864     & 0.865     & 0.077     & 0.091   & 0.012 & 0.048 & 0.04      & 0.042     & 0.117     & \textcolor{YellowOrange}{0.651} \\
                                & 5.0                          &                           & 25.659     & 0.840     & 0.094     & 0.072   & 0.008 & 0.039 & 0.025     & 0.037     & 0.108     & \textcolor{YellowOrange}{0.71}  \\ \cline{2-3}
                                & 1.0                          & \multirow{5}{*}{600}      & 29.626     & 0.910     & 0.049     & 0.191   & 0.058 & 0.088 & 0.085     & 0.084     & 0.129     & \textcolor{YellowOrange}{0.364} \\
                                & 2.0                          &                           & 26.599     & 0.861     & 0.083     & 0.123   & 0.01  & 0.059 & 0.042     & 0.052     & 0.119     & \textcolor{YellowOrange}{0.595} \\
                                & 3.0                          &                           & 24.665     & 0.819     & 0.109     & 0.078   & 0.006 & 0.043 & 0.025     & 0.032     & 0.097     & \textcolor{YellowOrange}{0.72}  \\
                                & 4.0                          &                           & 23.170     & 0.781     & 0.132     & 0.046   & 0.007 & 0.028 & 0.017     & 0.023     & 0.086     & \textcolor{YellowOrange}{0.794} \\
                                & 5.0                          &                           & 21.882     & 0.744     & 0.155     & 0.033   & 0.008 & 0.019 & 0.013     & 0.017     & 0.072     & \textcolor{YellowOrange}{0.838} \\ \hline
 \end{tabular}}
\caption{LDM-based emotion manipulation on the AffectNet validation set, using forward/backward DDIM processes with $T_{\textrm{DDIM}}=40$ steps, variable \textit{c.f.g.} scale $\gamma\in\{1.0,2.0,3.0,4.0,5.0\}$ and editing strength $t_{0}\in\{400,500,600\}$.}
\label{tab:baseline1}
\end{table*}

\begin{table*}[ht!]
\centering
\resizebox{11.6cm}{!}{
\begin{tabular}{c|c|c|n{3}{2}cc|ccccccc}
\hline
\multirow{2}{*}{$y_{\textrm{target}}$} & \multirow{2}{*}{DDIM Steps ($T_{\textrm{DDIM}}$)} & \multirow{2}{*}{Str. $(t_{0})$} & \multicolumn{3}{c|}{Image Quality} & \multicolumn{7}{c}{Emotion Recognition Accuracy w/ {HSEmotion} \cite{savchenko2021facial, savchenko2022video, savchenko2022classifying}}                 \\ \cline{4-13} 
                                &                             &                           & PSNR$\uparrow$       & SSIM$\uparrow$      & LPIPS$\downarrow$     & Neutral & Happy & Sad   & Surprised & Scared & Disgusted & Angry \\ \hline
\multirow{9}{*}{Happiness}      & 20                          & \multirow{3}{*}{400}      & 31.246     & 0.934     & 0.031     & 0.238   & \textcolor{YellowOrange}{0.451} & 0.028 & 0.15      & 0.035  & 0.033     & 0.065 \\
                                & 40                          &                           & 31.706     & 0.937     & 0.029     & 0.244   & \textcolor{YellowOrange}{0.447} & 0.028 & 0.148     & 0.035  & 0.032     & 0.066 \\
                                & 80                          &                           & 32.041     & 0.938     & 0.029     & 0.245   & \textcolor{YellowOrange}{0.446} & 0.03  & 0.15      & 0.034  & 0.031     & 0.065 \\ \cline{2-3}
                                & 20                          & \multirow{3}{*}{500}      & 27.667     & 0.883     & 0.064     & 0.133   & \textcolor{YellowOrange}{0.712} & 0.011 & 0.094     & 0.013  & 0.015     & 0.022 \\
                                & 40                          &                           & 27.923     & 0.883     & 0.065     & 0.13    & \textcolor{YellowOrange}{0.728} & 0.01  & 0.082     & 0.012  & 0.018     & 0.019 \\
                                & 80                          &                           & 28.177     & 0.883     & 0.065     & 0.127   & \textcolor{YellowOrange}{0.734} & 0.011 & 0.081     & 0.012  & 0.016     & 0.019 \\ \cline{2-3}
                                & 20                          & \multirow{3}{*}{600}      & 24.113     & 0.811     & 0.107     & 0.045   & \textcolor{YellowOrange}{0.905} & 0.002 & 0.031     & 0.004  & 0.007     & 0.005 \\
                                & 40                          &                           & 24.126     & 0.803     & 0.113     & 0.038   & \textcolor{YellowOrange}{0.916} & 0.003 & 0.027     & 0.004  & 0.008     & 0.004 \\
                                & 80                          &                           & 24.281     & 0.800     & 0.115     & 0.037   & \textcolor{YellowOrange}{0.919} & 0.003 & 0.026     & 0.004  & 0.007     & 0.005 \\ \hline
\multirow{9}{*}{Sadness}        & 20                          & \multirow{3}{*}{400}      & 31.184     & 0.933     & 0.031     & 0.157   & 0.052 & \textcolor{YellowOrange}{0.398} & 0.051     & 0.141  & 0.081     & 0.119 \\
                                & 40                          &                           & 31.676     & 0.937     & 0.029     & 0.159   & 0.052 & \textcolor{YellowOrange}{0.393} & 0.052     & 0.138  & 0.083     & 0.123 \\
                                & 80                          &                           & 32.024     & 0.938     & 0.028     & 0.159   & 0.054 & \textcolor{YellowOrange}{0.393} & 0.053     & 0.139  & 0.08      & 0.122 \\ \cline{2-3}
                                & 20                          & \multirow{3}{*}{500}      & 27.974     & 0.889     & 0.061     & 0.122   & 0.016 & \textcolor{YellowOrange}{0.562} & 0.029     & 0.111  & 0.055     & 0.104 \\
                                & 40                          &                           & 28.282     & 0.890     & 0.061     & 0.128   & 0.014 & \textcolor{YellowOrange}{0.573} & 0.027     & 0.106  & 0.052     & 0.1   \\
                                & 80                          &                           & 28.560     & 0.891     & 0.061     & 0.128   & 0.015 & \textcolor{YellowOrange}{0.572} & 0.027     & 0.105  & 0.052     & 0.102 \\ \cline{2-3}
                                & 20                          & \multirow{3}{*}{600}      & 24.995     & 0.833     & 0.097     & 0.089   & 0.006 & \textcolor{YellowOrange}{0.694} & 0.015     & 0.082  & 0.032     & 0.082 \\
                                & 40                          &                           & 25.062     & 0.827     & 0.103     & 0.09    & 0.008 & \textcolor{YellowOrange}{0.712} & 0.017     & 0.069  & 0.029     & 0.076 \\
                                & 80                          &                           & 25.231     & 0.824     & 0.106     & 0.089   & 0.008 & \textcolor{YellowOrange}{0.716} & 0.016     & 0.066  & 0.029     & 0.076 \\ \hline
\multirow{9}{*}{Surprise}       & 20                          & \multirow{3}{*}{400}      & 31.371     & 0.937     & 0.030     & 0.186   & 0.18  & 0.046 & \textcolor{YellowOrange}{0.34}      & 0.112  & 0.042     & 0.096 \\
                                & 40                          &                           & 31.873     & 0.940     & 0.028     & 0.191   & 0.18  & 0.048 & \textcolor{YellowOrange}{0.337}     & 0.11   & 0.04      & 0.094 \\
                                & 80                          &                           & 32.221     & 0.942     & 0.027     & 0.188   & 0.182 & 0.048 & \textcolor{YellowOrange}{0.339}     & 0.11   & 0.041     & 0.092 \\ \cline{2-3}
                                & 20                          & \multirow{3}{*}{500}      & 27.821     & 0.888     & 0.062     & 0.129   & 0.137 & 0.029 & \textcolor{YellowOrange}{0.512}     & 0.117  & 0.021     & 0.054 \\
                                & 40                          &                           & 28.096     & 0.889     & 0.063     & 0.127   & 0.136 & 0.027 & \textcolor{YellowOrange}{0.522}     & 0.115  & 0.024     & 0.048 \\
                                & 80                          &                           & 28.361     & 0.890     & 0.063     & 0.127   & 0.138 & 0.026 & \textcolor{YellowOrange}{0.525}     & 0.114  & 0.021     & 0.048 \\ \cline{2-3}
                                & 20                          & \multirow{3}{*}{600}      & 24.469     & 0.821     & 0.104     & 0.078   & 0.086 & 0.017 & \textcolor{YellowOrange}{0.661}     & 0.128  & 0.01      & 0.021 \\
                                & 40                          &                           & 24.448     & 0.813     & 0.110     & 0.078   & 0.085 & 0.016 & \textcolor{YellowOrange}{0.662}     & 0.133  & 0.008     & 0.018 \\
                                & 80                          &                           & 24.604     & 0.810     & 0.113     & 0.075   & 0.083 & 0.014 & \textcolor{YellowOrange}{0.67}      & 0.131  & 0.008     & 0.018 \\ \hline
\multirow{9}{*}{Fear}           & 20                          & \multirow{3}{*}{400}      & 30.986     & 0.933     & 0.031     & 0.084   & 0.068 & 0.126 & 0.144     & \textcolor{YellowOrange}{0.389}  & 0.08      & 0.108 \\
                                & 40                          &                           & 31.471     & 0.937     & 0.030     & 0.087   & 0.071 & 0.124 & 0.14      & \textcolor{YellowOrange}{0.388}  & 0.081     & 0.109 \\
                                & 80                          &                           & 31.817     & 0.938     & 0.029     & 0.087   & 0.072 & 0.121 & 0.142     & \textcolor{YellowOrange}{0.388}  & 0.08      & 0.11  \\ \cline{2-3}
                                & 20                          & \multirow{3}{*}{500}      & 27.457     & 0.883     & 0.065     & 0.043   & 0.034 & 0.098 & 0.14      & \textcolor{YellowOrange}{0.562}  & 0.053     & 0.072 \\
                                & 40                          &                           & 27.750     & 0.884     & 0.065     & 0.043   & 0.033 & 0.092 & 0.148     & \textcolor{YellowOrange}{0.562}  & 0.05      & 0.072 \\
                                & 80                          &                           & 28.034     & 0.885     & 0.065     & 0.042   & 0.032 & 0.094 & 0.146     & \textcolor{YellowOrange}{0.564}  & 0.05      & 0.072 \\ \cline{2-3}
                                & 20                          & \multirow{3}{*}{600}      & 24.092     & 0.813     & 0.109     & 0.018   & 0.016 & 0.068 & 0.133     & \textcolor{YellowOrange}{0.696}  & 0.027     & 0.041 \\
                                & 40                          &                           & 24.115     & 0.806     & 0.115     & 0.016   & 0.016 & 0.062 & 0.136     & \textcolor{YellowOrange}{0.708}  & 0.023     & 0.039 \\
                                & 80                          &                           & 24.286     & 0.804     & 0.117     & 0.016   & 0.016 & 0.061 & 0.134     & \textcolor{YellowOrange}{0.715}  & 0.021     & 0.038 \\ \hline
\multirow{9}{*}{Disgust}        & 20                          & \multirow{3}{*}{400}      & 31.091     & 0.934     & 0.031     & 0.116   & 0.072 & 0.099 & 0.074     & 0.11   & \textcolor{YellowOrange}{0.316}     & 0.214 \\
                                & 40                          &                           & 31.545     & 0.937     & 0.030     & 0.115   & 0.073 & 0.098 & 0.074     & 0.111  & \textcolor{YellowOrange}{0.317}     & 0.212 \\
                                & 80                          &                           & 31.881     & 0.938     & 0.029     & 0.115   & 0.073 & 0.097 & 0.077     & 0.108  & \textcolor{YellowOrange}{0.317}     & 0.213 \\ \cline{2-3}
                                & 20                          & \multirow{3}{*}{500}      & 27.916     & 0.890     & 0.061     & 0.079   & 0.041 & 0.071 & 0.04      & 0.079  & \textcolor{YellowOrange}{0.466}     & 0.224 \\
                                & 40                          &                           & 28.214     & 0.891     & 0.062     & 0.073   & 0.04  & 0.072 & 0.04      & 0.073  & \textcolor{YellowOrange}{0.478}     & 0.224 \\
                                & 80                          &                           & 28.480     & 0.891     & 0.062     & 0.074   & 0.041 & 0.072 & 0.04      & 0.072  & \textcolor{YellowOrange}{0.481}     & 0.219 \\ \cline{2-3}
                                & 20                          & \multirow{3}{*}{600}      & 24.938     & 0.833     & 0.100     & 0.041   & 0.032 & 0.05  & 0.025     & 0.051  & \textcolor{YellowOrange}{0.579}     & 0.222 \\
                                & 40                          &                           & 24.981     & 0.826     & 0.106     & 0.038   & 0.033 & 0.047 & 0.02      & 0.052  & \textcolor{YellowOrange}{0.593}     & 0.216 \\
                                & 80                          &                           & 25.154     & 0.824     & 0.108     & 0.038   & 0.036 & 0.05  & 0.019     & 0.048  & \textcolor{YellowOrange}{0.596}     & 0.212 \\ \hline
\multirow{9}{*}{Anger}          & 20                          & \multirow{3}{*}{400}      & 31.424     & 0.937     & 0.029     & 0.169   & 0.069 & 0.083 & 0.082     & 0.088  & 0.131     & \textcolor{YellowOrange}{0.377} \\
                                & 40                          &                           & 31.943     & 0.940     & 0.027     & 0.168   & 0.071 & 0.085 & 0.084     & 0.088  & 0.132     & \textcolor{YellowOrange}{0.372} \\
                                & 80                          &                           & 32.298     & 0.942     & 0.026     & 0.17    & 0.07  & 0.083 & 0.083     & 0.088  & 0.133     & \textcolor{YellowOrange}{0.374} \\ \cline{2-3}
                                & 20                          & \multirow{3}{*}{500}      & 27.978     & 0.890     & 0.060     & 0.125   & 0.02  & 0.064 & 0.053     & 0.055  & 0.128     & \textcolor{YellowOrange}{0.554} \\
                                & 40                          &                           & 28.304     & 0.891     & 0.061     & 0.127   & 0.021 & 0.063 & 0.05      & 0.054  & 0.128     & \textcolor{YellowOrange}{0.558} \\
                                & 80                          &                           & 28.587     & 0.892     & 0.061     & 0.13    & 0.02  & 0.061 & 0.051     & 0.052  & 0.125     & \textcolor{YellowOrange}{0.562} \\ \cline{2-3}
                                & 20                          & \multirow{3}{*}{600}      & 24.599     & 0.825     & 0.103     & 0.078   & 0.006 & 0.043 & 0.026     & 0.033  & 0.105     & \textcolor{YellowOrange}{0.709} \\
                                & 40                          &                           & 24.665     & 0.819     & 0.109     & 0.078   & 0.006 & 0.043 & 0.025     & 0.032  & 0.097     & \textcolor{YellowOrange}{0.72}  \\
                                & 80                          &                           & 24.832     & 0.817     & 0.112     & 0.075   & 0.005 & 0.041 & 0.024     & 0.029  & 0.1       & \textcolor{YellowOrange}{0.725} \\ \hline
\end{tabular}}
\caption{LDM-based emotion manipulation on the AffectNet validation set, using forward/backward DDIM processes with \textit{c.f.g.} scale $\gamma=3.0$, variable number of steps $T_{\textrm{DDIM}}\in\{20,40,80\}$ and editing strength $t_{0}\in\{400,500,600\}$.}
\label{tab:baseline2}
\end{table*}

\section{CLIP-Guided Finetuning}

The first step in finetuning an LDM, involves the precomputation of noisy latents for a predefined number of training instances and the entirety of the testing instances. We precomputed noisy latents using deterministic forward DDIM ($\eta=0$) with $T_{\textrm{DDIM}}=40$ steps and $t_{0}=500$ editing strength for 4,000 training instances (500 per emotion class). Subsequently, we randomly sampled 1,000 latents from the latter pool in an attempt to trade off model performance against reduced tuning time. This procedure was replicated for all guidance scales $\gamma\in\{1.0,2.0,3.0,4.0,5.0\}$. We ran finetuning for 20 epochs with a base learning rate of $2\times10^{-6}$, batch size of 4, using the AdamW \cite{loshchilov2018decoupled} optimizer. Each batch of latents undergoes DDIM sampling for $T_{\textrm{tune}}=6$ steps and $t_{0}=500$. Obviously this low number of steps in the DDIM sampling process gives imperfect reconstruction but is empirically chosen \cite{kim2022diffusionclip} as an acceptable compromise between sample quality and GPU memory requirements. The tuning procedure is repeated for each one of the six target emotions, leading to six different models per value of $\gamma$ and combination of $(\lambda_{\textrm{dir}},\lambda_{\textrm{id}},\lambda_{\ell_{2}})$ coefficients. For simplicity, during all experiments we set $\lambda_{\textrm{id}}=\lambda_{\ell_{2}}=1$ and only varied $\lambda_{\textrm{dir}}$.

\begin{figure*}[t!]
\centering
\begin{tabular}{cc}
\subfloat[Baseline]{\includegraphics[height=3.2in]{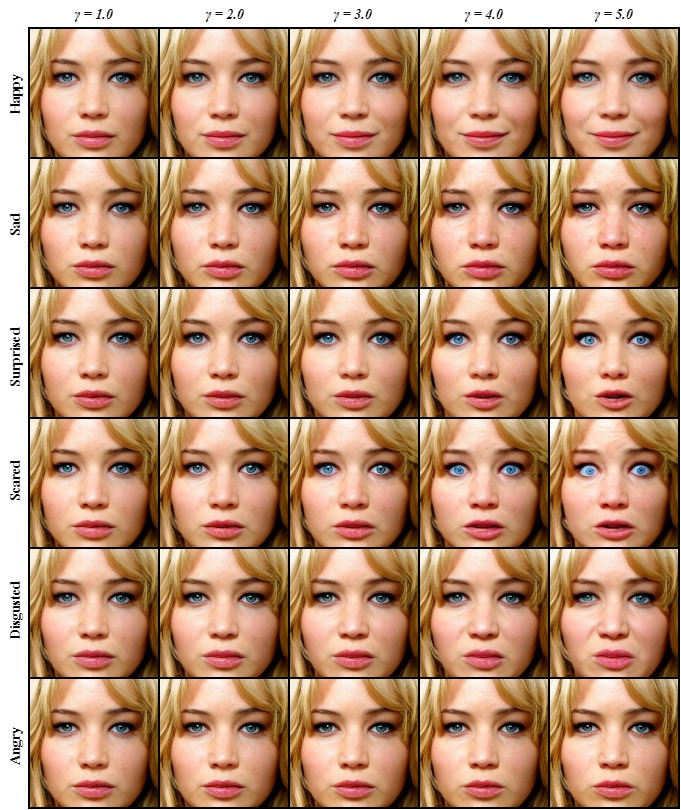}} &
\subfloat[Finetuned]{\includegraphics[height=3.2in]{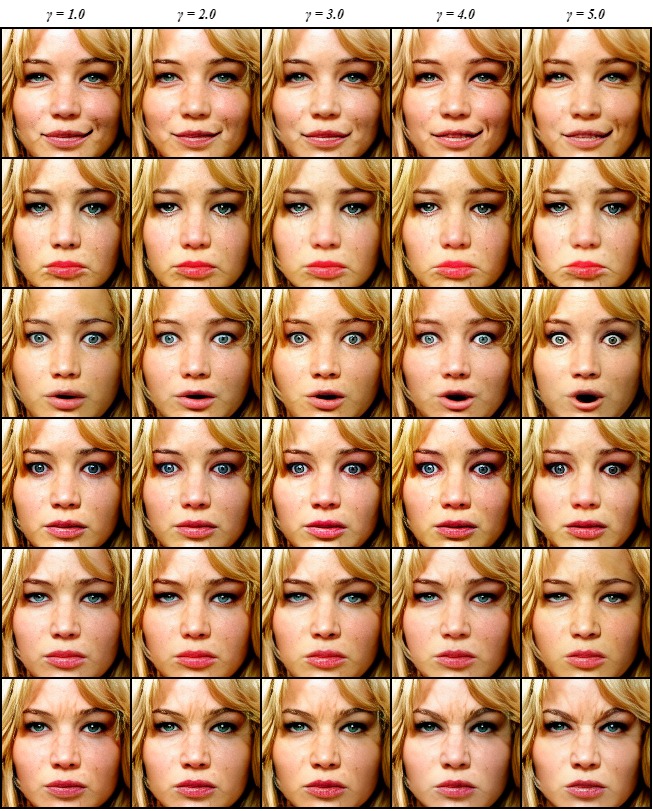}} \\ 
\end{tabular}
\caption{Emotion translation comparison between baseline and finetuned models on curated examples from the AffectNet validation set, using $T_{\textrm{DDIM}}=40$ steps, $\eta=0$, editing strength $t_{0}=500$, variable \textit{c.f.g.} scale $\gamma\in\{1.0,2.0,3.0,4.0,5.0\}$, $\lambda_{\textrm{dir}}=2.0, \lambda_{\textrm{id}}=\lambda_{\ell_{2}}=1.0$.}
\label{fig:baseline_vs_tuned_var_scale}
\end{figure*}

\begin{figure*}[t!]
\centering
\begin{tabular}{cc}
\subfloat{\includegraphics[height=1.85in]{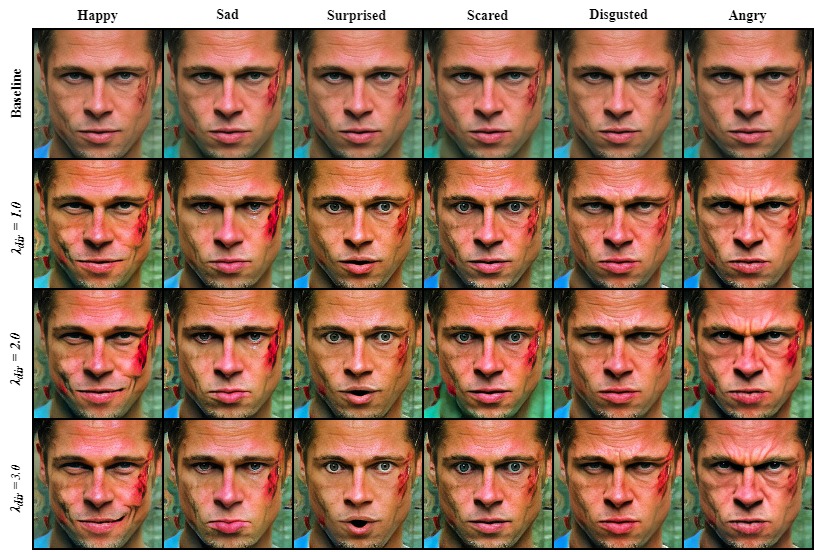}} &
\subfloat{\includegraphics[height=1.85in]{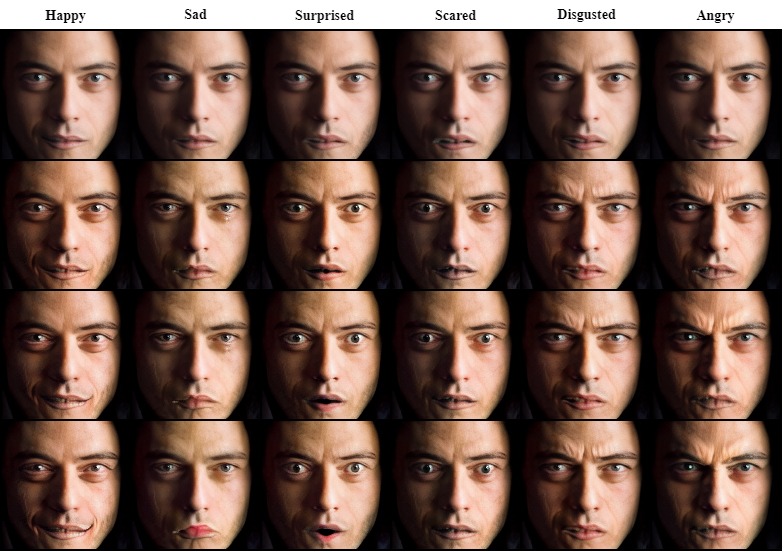}} \\ 
\end{tabular}
\caption{Qualitative comparison between baseline and finetuned models on curated examples from the AffectNet validation set, using $T_{\textrm{DDIM}}=40$ steps, $\eta=0$, editing strength $t_{0}=500$, $\gamma=1$  $\lambda_{\textrm{id}}=\lambda_{\ell_{2}}=1.0$ and variable $\lambda_{\textrm{dir}}\in\{1.0,2.0,3.0\}$.}
\label{fig:baseline_vs_tuned_var_lambda}
\end{figure*}

A qualitative comparison between the baseline and tuned models is presented in \cref{fig:baseline_vs_tuned_var_scale}, \cref{fig:baseline_vs_tuned_var_lambda}, where the values of \textit{c.f.g} scale $\gamma$ and weight $\lambda_{\textrm{dir}}$ are varied, respectively. Increasing $\gamma$ in the finetuned models, does not particularly affect the cases of happiness and disgust while a slight shift in colors can be noticed in some cases. This could either be a result of the small number of sampling steps used during the tuning process ($T_{\textrm{tune}}=6$) or higher required coefficients for $\lambda_{\ell_{2}}$. 

\begin{figure*}[t!]
    \centering
    \begin{minipage}{0.75\textwidth}
        \centering
        \includegraphics[width=1.0\textwidth]{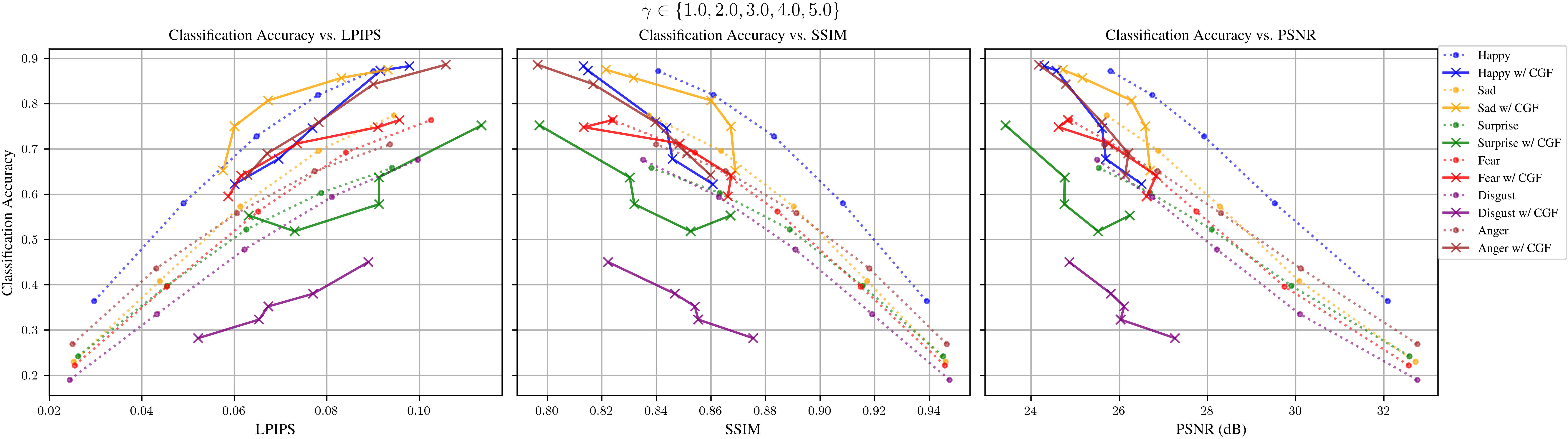} 
    \end{minipage}%
    \caption{Quantitative comparison between baseline and CLIP-guided finetuned (CGF) models, in terms of mean classification accuracy (in the range [0-1]), LPIPS, SSIM and PSNR of manipulated samples, using $T_{\textrm{DDIM}}=40$ steps, editing strength $t_{0}=500$, $\lambda_{\textrm{dir}}=2.0, \lambda_{\textrm{id}}=\lambda_{\ell_{2}}=1.0$, across all \textit{c.f.g.} scales $\gamma\in\{1.0,2.0,3.0,4.0,5.0\}$.}
    \label{fig:baseline_vs_tuned_metrics_var_scale}
\end{figure*}

\begin{figure*}[t!]
    \centering
    \begin{minipage}{0.75\textwidth}
        \centering
        \includegraphics[width=1.0\textwidth]{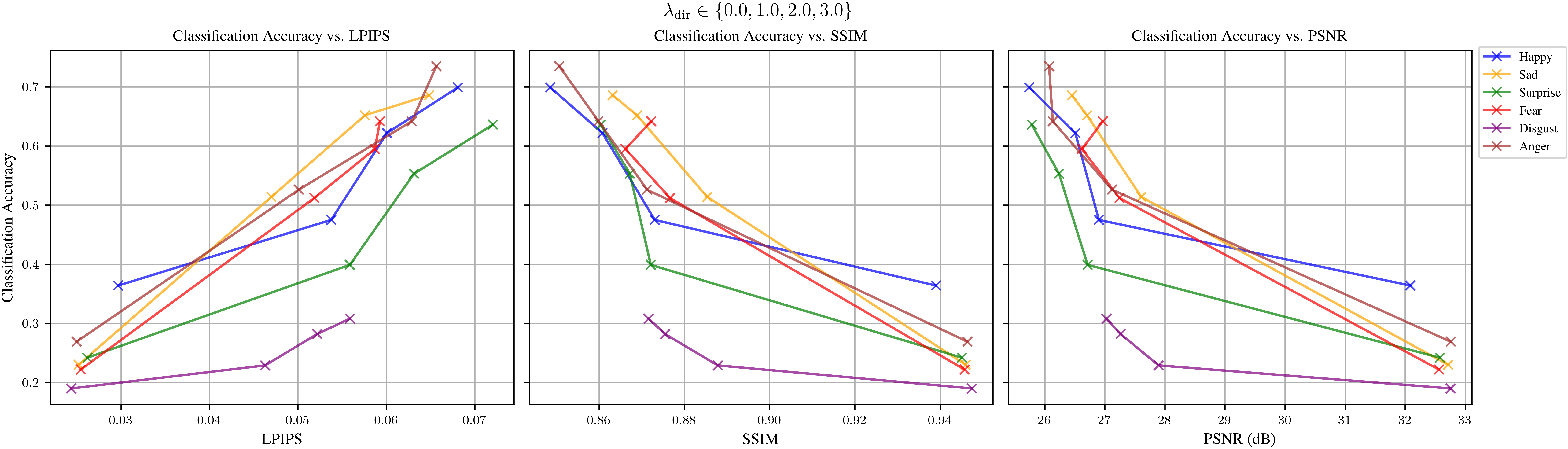} 
    \end{minipage}%
    \caption{Quantitative comparison between baseline $(\lambda_{\textrm{dir}}=0)$ and CLIP-guided finetuned (CGF) models $(\lambda_{\textrm{dir}}>0)$, in terms of mean classification accuracy (in the range [0-1]), LPIPS, SSIM and PSNR of manipulated samples, using $T_{\textrm{DDIM}}=40$ steps, editing strength $t_{0}=500$, $\gamma=1.0, \lambda_{\textrm{id}}=\lambda_{\ell_{2}}=1.0$, across different values of  $\lambda_{\textrm{dir}}\in\{0.0,1.0,2.0,3.0\}$.}
    \label{fig:baseline_vs_tuned_metrics_var_lambda}
\end{figure*}

\begin{figure}[t!]
    \centering
    \begin{minipage}{0.4\textwidth}
        \centering
        \includegraphics[width=1.0\textwidth]{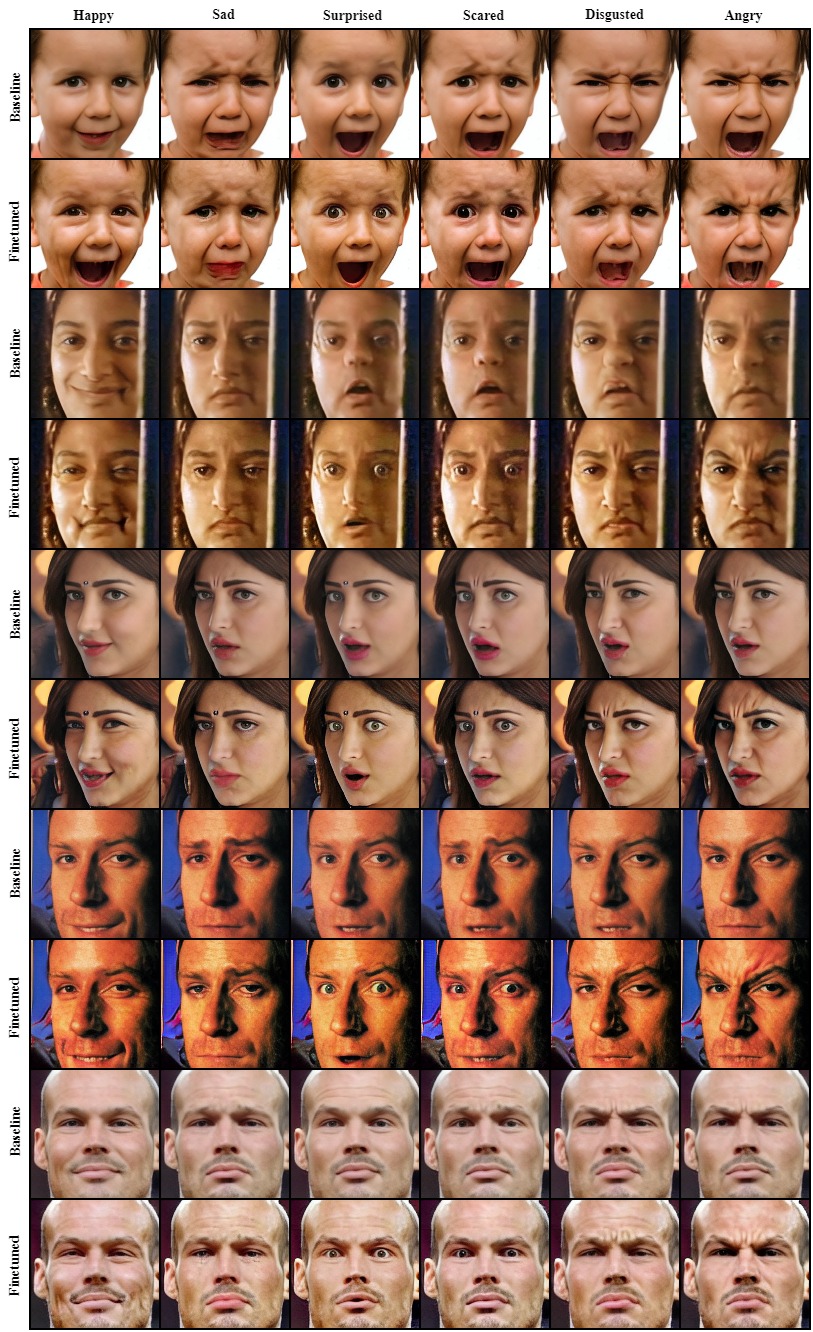} 
    \end{minipage}%
    \caption{Qualitative comparison between baseline and finetuned models on curated examples from the AffectNet validation set, using $T_{\textrm{DDIM}}=40$ steps, $\eta=0$, editing strength $t_{0}=500$, \textit{c.f.g} scale $\gamma=3.0$, $\lambda_{\textrm{dir}}=2.0, \lambda_{\textrm{id}}=\lambda_{\ell_{2}}=1.0$.}
    \label{fig:baseline_vs_tuned_var_sample}
\end{figure}

\cref{fig:baseline_vs_tuned_metrics_var_scale} and \cref{fig:baseline_vs_tuned_metrics_var_lambda} present additional quantitative comparisons between baseline and finetuned models, as well as insights regarding the trade-off between emotion translation and quality preservation during CGF. More specifically, in \cref{fig:baseline_vs_tuned_metrics_var_scale} the finetuned models demonstrate higher classification accuracy scores for the same values of \textit{c.f.g} scale $\gamma$, with the exception of the emotion of `disgust'. Images that undergo CLIP guidance with the latter being the target emotion, end up looking very similar to their `angry' counterparts, hence the drop in accuracy. The boost in classification accuracy comes at the cost of a slight dicrease in image quality across all quality metrics, but as it will be demonstrated later, the end result still remains visually compelling. Furthermore, in \cref{fig:baseline_vs_tuned_metrics_var_lambda}, we can see that even with no unconditional guidance ($\gamma=1$), a setting in which the manipulation effect of the baseline method is often unnoticeable to the human eye, the introduction of directional CLIP loss resolves emotional ambiguity to a large extend, with the exception of `disgust', as discussed above.  An additional qualitative comparison between baseline and finetuned models on curated examples can be seen in \cref{fig:baseline_vs_tuned_var_sample}.

\section{Training \& Evaluation Specifications}
Code regarding the implementation of our models and reproduction of all experiments is included as supplementary material. The backbone of our code is based on the original Stable Diffusion repository\footnote{\url{https://github.com/CompVis/stable-diffusion}}.

For testing GANmut \cite{d2021ganmut} against our own models, we used the official codebase\footnote{\url{https://github.com/stefanodapolito/GANmut}} along with publicly available models that have been pre-trained on AffectNet. We evaluated both linear (GANmut) and Gaussian (GGANmut) models. In the case of the linear model, we manipulated images using maximum intensity $(\rho=1)$. In order to achieve that, we use the learned unit vectors $\bm{v}_{c}=(x_{c},y_{c})$ corresponding to each emotion $c\in\{1,\dots,C\}$, and calculate the principal direction $\theta_{c}=\textrm{arctan2}(y_{c},x_{c})$. 

Subsequently, for the application of GANimation \cite{pumarola2018ganimation} in the emotion manipulation scenario with discrete emotions, we used the following scheme: we randomly sampled 500 images corresponding to each emotion label from the training set and computed the required target AU vectors for each one of them, using the publicly available OpenFace\footnote{\url{https://github.com/TadasBaltrusaitis/OpenFace}} \cite{baltrusaitis2018openface} toolkit. Each vector contains intensities in the range $[0,5]$ for each on of 17 detected AUs which are later normalized to $[0,1]$. As GAN training is notoriously unstable, we opted for using a publicly available model checkpoint\footnote{\mbox{\url{https://github.com/donydchen/ganimation_replicate}}}, pre-trained on more than 400K images with annotated AUs belonging to the EmotionNet database \cite{fabian2016emotionet}. 

We trained a StarGAN v2 \cite{choi2020stargan} model, using the publicly available official codebase\footnote{\url{https://github.com/clovaai/stargan-v2}}, on AffectNet for 100K iterations using a batch size of 8. The learning rate for $D,E$ and $G$ is set equal to $10^{-4}$, while for $F$ is set equal to $10^{-6}$, using the Adam optimizer. The loss weighting coefficients were set equal to $\lambda_{\textrm{sty}}=\lambda_{\textrm{div}}=\lambda_{\textrm{cyc}}=\lambda_{\textrm{adv}}=1$, for simplicity. Input image size is $128^{2}$ pixels, with $|\mathcal{Y}|=8$ domains (in accordance with the basic 7 emotions of the {AffectNet} dataset, plus neutral). We used latent code dimension $d_{\bm{z}}=16$, style code dimension $d_{\bm{s}}=64$ and 512 units for the hidden layers of the mapping network. For image-manipulation, we used latent-based translation, generating 10K latent codes per image, per target emotion. Subsequently, these latent codes were mapped to their corresponding style codes using the trained mapping network, while the generator was fed with the average of the computed style codes, along with the images to be translated. 

\section{Comparisons w/ GAN-based Models}
\begin{figure*}[t!]
\centering
\begin{tabular}{ccc}
\subfloat{\includegraphics[height=7.2in]{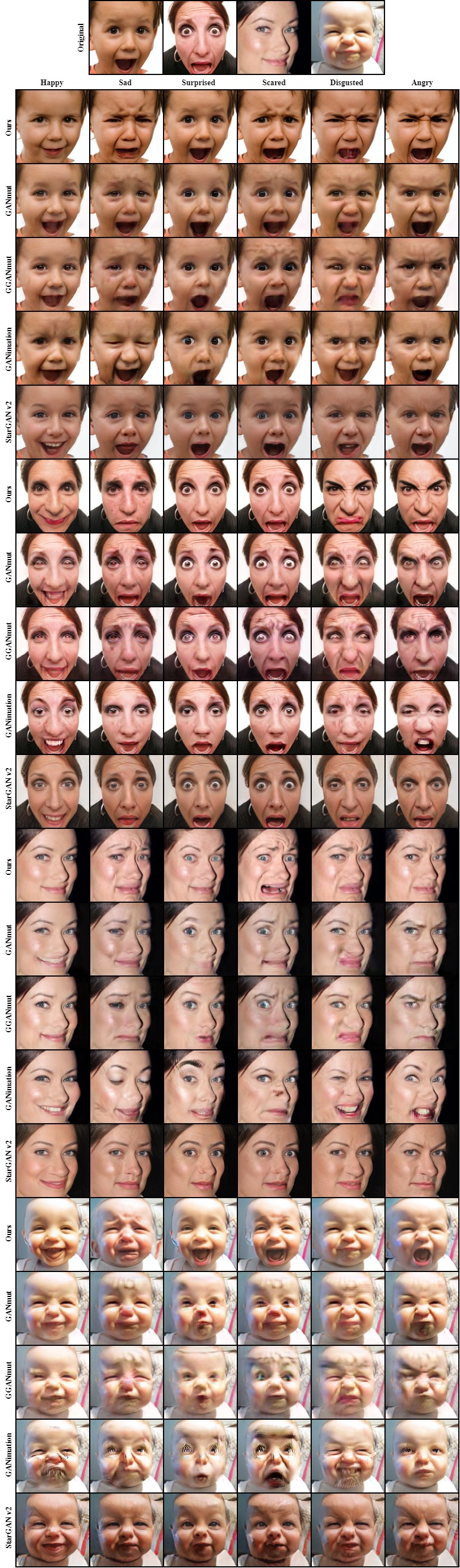}} &
\subfloat{\includegraphics[height=7.2in]{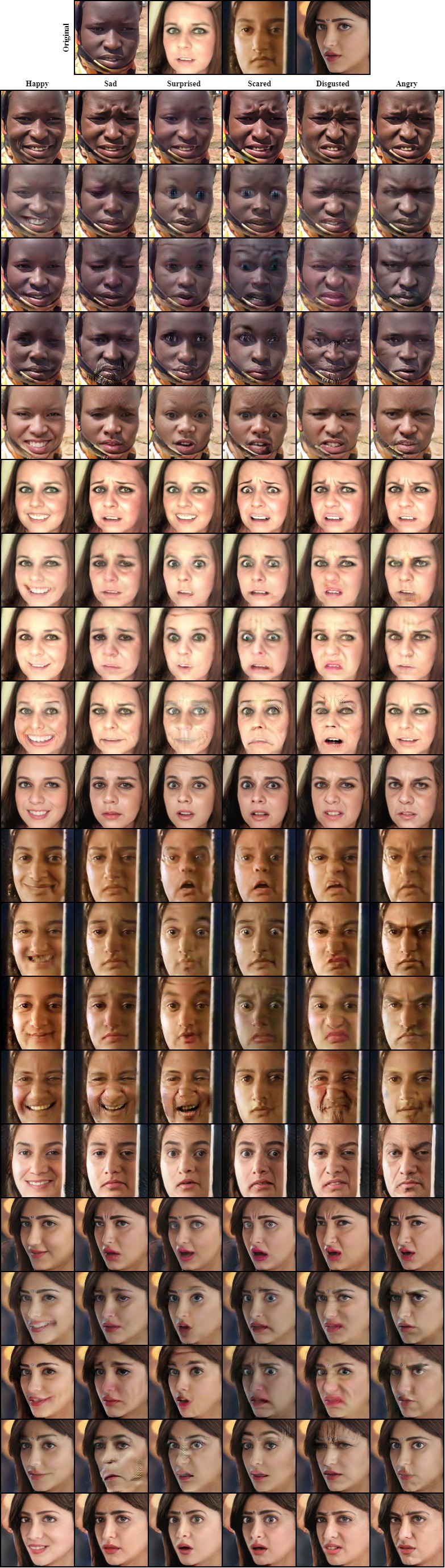}} &
\subfloat{\includegraphics[height=7.2in]{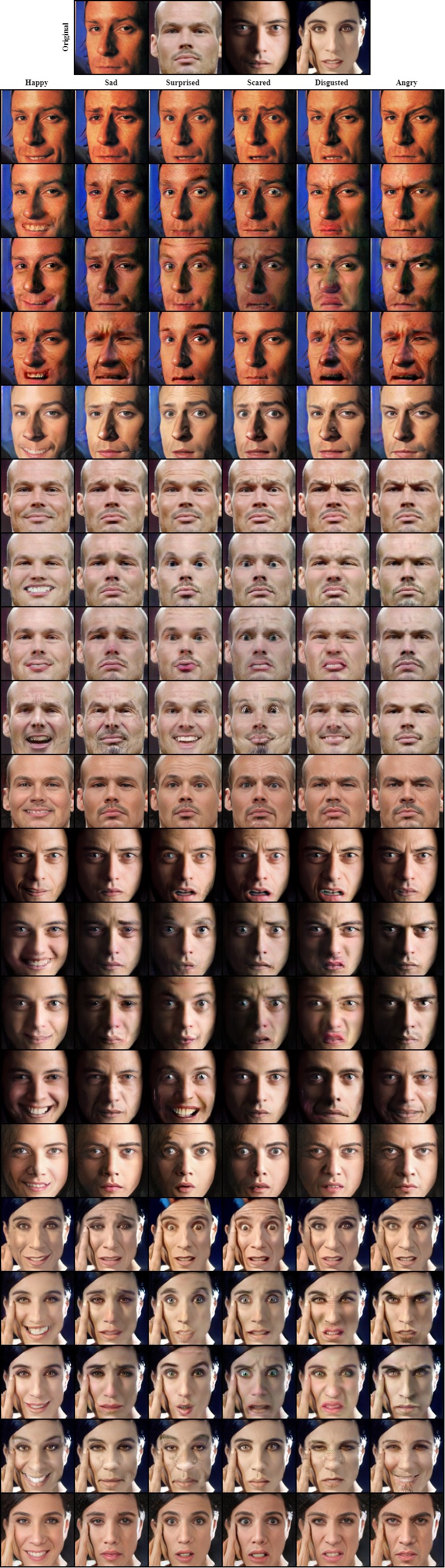}} \\ 
\end{tabular}
\caption{Qualitative comparison among our LDM-based model and GAN-based implementations for image-based emotional manipulation on curated examples from the AffectNet validation set.}
\label{fig:comparisons_a}
\end{figure*}

\cref{fig:comparisons_a} provides a qualitative comparison among the three aforementioned implementations, considering both \textit{linear} and \textit{Gaussian} variants of GANmut, on curated examples from the AffectNet validation set. Making comparisons between GAN and diffusion-based models is far from straightforward and we need to take into consideration the underlying trade-off between emotion classification accuracy and quality/identity preservation with regard to source images. We immediately notice that samples manipulated with GANmut, feature accurate depictions of the target emotions but often suffer from noise and subject identity deformation. Moreover, as it was expected, it is far more difficult to capture the desired emotions with GANimation and randomly selected driving AUs, rather that curated ones, but such a comparison would have been unfair with regard to other implementations. On the other hand, StarGAN v2 seems to strike a better balance between image quality and emotion transfer, but mostly fails to preserve the subjects' identity, skin tone and illumination. Our diffusion model successfully compromises between style preservation and emotion transfer, while also maintaining high image quality.

\begin{table}[t!]
\centering
\resizebox{8cm}{!}{
\begin{tabular}{c|ccccc}
\multirow{2}{*}{Method} & \multicolumn{5}{c}{Mean}                                                    \\ \cline{2-6} 
                        & Accuracy$\uparrow$ & PSNR$\uparrow$           & SSIM$\uparrow$           & LPIPS$\downarrow$          & CSIM$\uparrow$           \\ \hline
Groundtruth \cite{mollahosseini2017affectnet}              & 0.638    & --          & --          & --          & --          \\
GANimation \cite{pumarola2018ganimation}              & 0.348    & 24.17          & 0.822          & 0.100          & 0.562          \\
StarGAN v2 \cite{choi2020stargan}              & 0.782    & 17.95          & 0.684          & 0.157          & 0.514          \\
GANmut \cite{d2021ganmut}                 & 0.877    & 22.55          & 0.824          & 0.106          & 0.693          \\
GGANmut \cite{d2021ganmut}                 & \textbf{0.969}    & 21.67          & 0.783          & 0.123          & 0.614          \\
Ours                   & 0.742    & \textbf{25.51} & \textbf{0.836} & \textbf{0.096} & 0.738          \\
Ours w/ CGF             & 0.768    & 24.38          & 0.813          & 0.099          & \textbf{0.745} \\ \hline
\end{tabular}}
\caption{Quantitative comparison with mean aggregated evaluation metrics across all six target emotions.}
\label{tab:mean_comparisons}
\end{table}

\cref{tab:mean_comparisons} showcases mean aggregated evaluation measures in terms both image quality and emotion translation accuracy across all six of the considered target emotions, between our models (baseline and w/ CGF) and the aforementioned GAN-based methodologies.

\end{document}